\documentclass{article}

\usepackage[dblblindworkshop, final]{neurips_2026}
\workshoptitle{Geometric Distributional Deep Learning: Bridging Optimal Transport, Learning and Structured Data}

\usepackage[utf8]{inputenc} 
\usepackage[T1]{fontenc}    
\usepackage{hyperref}       
\usepackage{url}            
\usepackage{booktabs}       
\usepackage{amsfonts}       
\usepackage{nicefrac}       
\usepackage{microtype}      
\usepackage{xcolor}         

\usepackage{amsmath,amssymb,amsthm}
\usepackage{booktabs}
\usepackage{graphicx}
\usepackage{xcolor}
\usepackage{geometry}
\usepackage{multirow}

\newcommand{\rstar}{r^{\star}}
\newcommand{\ptgt}{\pi}
\newcommand{\nsrc}{n^{\mathrm{s}}}
\newcommand{\ntgt}{n^{\mathrm{t}}}
\newcommand{\Tstar}{T^{\star}}

\title{Disentangling Marginal Capacity from Geometry in Optimal-Transport Label Transfer}

\author{%
\begin{tabular}{ccc}
Ali Khodabandeh Yalabadi$^{1}$\thanks{Equal contribution}
& Azim Dehghani Amirabad$^{1}$\footnotemark[1]
& Artem Moskalev$^{2}$
\end{tabular}
\\[0.4em]
\bfseries
\begin{tabular}{cccc}
Yun-Ching Chen$^{1}$
& Mehdi Pirooznia$^{1}$
& Scott Oloff$^{1}$
& Rui Liao$^{1}$\thanks{Corresponding author: Rui Liao (RLiao2@its.jnj.com)}
\end{tabular}
\\[0.8em]
$^{1}$Johnson \& Johnson Innovative Medicine \qquad
$^{2}$Independent Researcher
}

\begin{document}

\maketitle

\begin{abstract}
Optimal transport (OT) is widely used to align distributions and transfer labels between unpaired source and target data, but source–target composition shifts can cause rare classes to be systematically under-predicted even with high transport confidence. We identify a source of this failure in the transport marginal: under the standard uniform weighting of source points, each class has transport capacity proportional to its source frequency, which can be insufficient to meet its target demand even when separated well in the embedding space. We show that displaced mass is preferentially absorbed by high-capacity classes rather than geometrically similar classes, producing confident errors that are difficult to detect from the transport plan. Downstream read-outs, entropic smoothing, and marginal relaxation provide limited recovery, whereas matching the transport marginal to the target composition substantially improves label transfer. On single-cell data, the oracle composition-matched marginal raises macro-F1 from 0.322 to 0.623, and the best deployable estimator closes 69\% of the oracle gap. Correcting the marginal also separates a coverage-dependent component of error from residual error associated with class separability, showing that transport capacity and representation geometry are distinct sources of transfer error.
\end{abstract}

\section{Introduction}
\label{sec:intro}

\begin{figure}[t!]
\centering
\includegraphics[width=\textwidth]{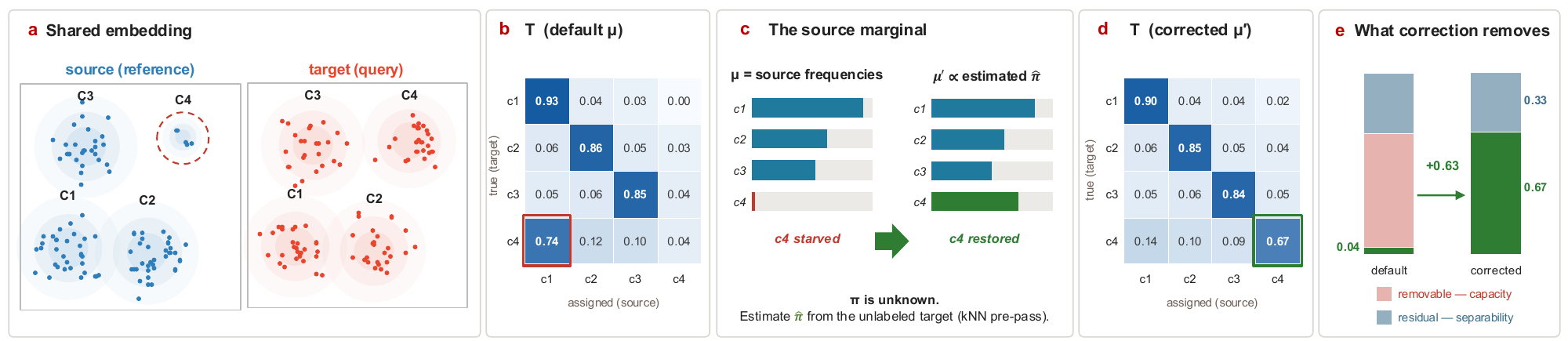}
\caption{\textbf{Capacity and correction.}
\textbf{(a)} $c_4$ is rare in the source yet well separated; $c_1,c_2$ overlap.
\textbf{(b)} Under the default $\mu=$ source frequencies, $c_4$ is starved and its mass is absorbed by the abundant $c_1$.
\textbf{(c)} Replacing it with $\mu'=\hat{\pi}/n^{\mathrm{s}}$, $\hat{\pi}$ estimating the target composition, restores $c_4$'s budget.
\textbf{(d)} Correct assignments return; the $c_1\!\leftrightarrow\!c_2$ off-diagonal is unchanged.
\textbf{(e)} Recovered mass is the capacity term; the residual tracks separability.}
\label{fig:capacity}
\vspace{-2 mm}
\end{figure}

Optimal transport (OT) is widely used to transfer labels between unpaired source and target data: given a shared representation, target labels are inferred from the source mass transported to each target point. This formulation appears across domain adaptation \citep{courty2017optimal,fatras2021unbalanced} and cross-domain alignment \citep{cao2022uniport,cao2022glue,klein2025mapping,bunne2024optimaltransport}, where the transport coupling provides a common mechanism for matching source and target distributions. Yet when source and target compositions differ, rare classes can transfer poorly despite high confidence, with errors surfacing as false positives on abundant classes (Fig.~\ref{fig:capacity}). This pattern is often attributed to representation quality \citep{tejada2025nicheformer,cui2024scgpt,kedzierska2025zeroshot}, but can the transport objective itself create this asymmetry?

We study this question through the source marginal of the transport plan. The marginal determines how much transport mass each source class can supply, and under the standard uniform weighting of source points \citep{cuturi2013sinkhorn,peyre2019computational}, this capacity is proportional to source class frequency. When the target composition differs from the source composition, a class may therefore have less transport mass available than the target mass it needs, \textit{even when the embedding separates it well}.

This failure is difficult to detect because rare classes can be displaced into large, well-supported classes with high transport confidence (Fig.~\ref{fig:capacity}b). Because the displaced mass can still receive high-confidence labels, output-side statistics can obscure rather than reveal the underlying capacity failure (Sec.~\ref{sec:mechanism}). Downstream modifications of the transport plan are similarly limited: alternative read-outs \citep{el2021label}, entropic smoothing \citep{cuturi2013sinkhorn}, and relaxing the marginal constraint recover only a fraction of the lost rare-class performance, while post-hoc class-prior reweighting recovers substantially less than correcting the marginal itself (Sec.~\ref{sec:knobs}). This limitation is a consequence of where the failure occurs: it arises in the transport coupling, before the final label read-out.

We therefore reshape the transport marginal to reflect the target composition rather than the source frequencies. In practice, however, the target composition is unknown, and we test whether it can be estimated from unlabeled target data well enough to recover the lost rare-class performance (Fig.~\ref{fig:capacity}c, Sec.~\ref{sec:correction}). We evaluate this composition-matched marginal when source and target compositions differ under two complementary domain shifts: cross-modal single-cell transfer \citep{janesick2023xenium,li2022benchmarking}, where the measurement process changes, and image domain adaptation \citep{venkateswara2017deep}, where the shift is primarily stylistic and coverage can be controlled independently of class separability. The oracle composition-matched marginal raises macro-F1 for OT label transfer from 0.322 to 0.623 on single-cell data, while the best deployable estimator closes 69\% of the oracle gap on single-cell data and 59\% on Office-Home (Sec.~\ref{sec:correction}). We also find that the correction should be applied selectively: its benefit is concentrated in capacity-limited classes and can become negative when coverage is sufficient (Sec.~\ref{sec:correction}). Together, these results show that correcting the transport marginal removes a coverage-dependent component of label-transfer error and exposes a residual component associated with the representation.

To summarize, we make the following contributions:

\begin{itemize}\itemsep1pt
\item \textbf{Capacity-driven failure.}
We identify marginal capacity as a source of rare-class transfer errors, with displaced mass flowing to high-capacity rather than geometrically similar classes.

\item \textbf{Upstream failure.}
We show that these errors are difficult to detect and that downstream read-outs, smoothing, and marginal relaxation provide limited recovery.

\item \textbf{Marginal correction.}
Matching the marginal to the target composition substantially improves transfer, but only for capacity-limited classes; its practical use depends on estimating the target composition.

\item \textbf{Capacity--geometry separation.}
Correcting the marginal removes the coverage-dependent component of error, revealing residual error associated with class separability.
\end{itemize}

\section{Optimal-transport label transfer and marginal capacity}
\label{sec:setup}

\paragraph{Optimal-transport label transfer}
Let $\mathcal{D}^{\mathrm{s}}=\{(x^{\mathrm{s}}_j,y_j)\}_{j=1}^{\nsrc}$ be a labelled source dataset of $\nsrc$ points, with $x^{\mathrm{s}}_j\in\mathcal{X}^{\mathrm{s}}$ and class labels $y_j\in\{1,\ldots,K\}$, and let $\mathcal{D}^{\mathrm{t}}=\{x^{\mathrm{t}}_i\}_{i=1}^{\ntgt}$ be an unlabelled target dataset of $\ntgt$ points, with $x^{\mathrm{t}}_i\in\mathcal{X}^{\mathrm{t}}$. A frozen encoder $\phi\colon\mathcal{X}^{\mathrm{s}}\cup\mathcal{X}^{\mathrm{t}}\to\mathbb{R}^{d}$ maps source and target data into a shared $d$-dimensional space. We use the squared Euclidean distance there as the transport cost \citep{peyre2019computational,courty2017optimal},

\[
C_{ij}
=
\left\lVert
\phi(x^{\mathrm{t}}_i)-\phi(x^{\mathrm{s}}_j)
\right\rVert^{2}.
\]

Let $\mu\in\Delta^{\nsrc}$ and $\nu\in\Delta^{\ntgt}$ denote the source and target marginals, where $\Delta^n$ is the probability simplex in $\mathbb{R}^{n}$. We use a uniform target marginal throughout, $\nu_i=1/\ntgt$. The transport plan $T\in\mathbb{R}_{+}^{\ntgt\times\nsrc}$ is obtained by solving entropic unbalanced OT (UOT) \citep{cuturi2013sinkhorn,chizat2018scaling,sejourne2023unbalanced}:

\begin{equation}
\Tstar
=
\operatorname*{arg\,min}_{T\ge0}\;
\langle C,T\rangle
+\varepsilon H(T)
+\tau\!\left[
\mathrm{KL}\!\left(T\mathbf{1}\,\middle\|\,\nu\right)
+\mathrm{KL}\!\left(T^{\!\top}\mathbf{1}\,\middle\|\,\mu\right)
\right].
\label{eq:uot}
\end{equation}

Here $\langle C,T\rangle=\sum_{ij}C_{ij}T_{ij}$, $H(T)=\sum_{ij}T_{ij}(\log T_{ij}-1)$, and $\mathrm{KL}(a\,\|\,b)=\sum_{\ell}[a_\ell\log(a_\ell/b_\ell)-a_\ell+b_\ell]$ is the generalized Kullback--Leibler divergence. The parameter $\varepsilon$ controls the entropic regularization, while $\tau$ controls how strongly the source and target marginals are enforced; as $\tau\to\infty$, the solution approaches balanced OT.

\paragraph{Marginal capacity} The source marginal $\mu$ determines the nominal amount of transport mass available from each source class. We refer to this class-level mass as its \emph{marginal capacity}: class $k$ has marginal capacity $\sum_{j:y_j=k}\mu_j$. Under the standard choice $\mu_j=1/\nsrc$ \citep{cuturi2013sinkhorn,peyre2019computational}, this becomes $\nsrc_k/\nsrc$, where $\nsrc_k$ is the number of source points of class $k$. Thus, marginal capacity is proportional to source class frequency.

\paragraph{Read-outs} The transport plan is converted into target labels by aggregating transported mass from source points of each class. The standard read-out is
\[
\hat{y}_i
=
\operatorname*{arg\,max}_{k}
\sum_{j:\,y_j=k}\Tstar_{ij}.
\]
Other read-outs use the same class-aggregated mass, including direct or tempered posterior probabilities and label propagation \citep{el2021label}. Because these read-outs operate on a fixed transport plan, they can be varied independently of the transport solve; we use this separation in Sec.~\ref{sec:failure}.

\paragraph{Coverage ratio}
For each class $k$, let $\ntgt_k$ denote the number of target points of that class and let $\pi_k=\ntgt_k/\ntgt$ denote its target composition. We measure source supply relative to target demand using the coverage ratio
\[
r_k=\frac{\nsrc_k}{\ntgt_k},
\]
the number of source points of class $k$ available per target point of that class. Under uniform $\nu$, the target marginal assigns total mass $\pi_k$ to the target points of class $k$.

\section{The Capacity Failure}
\label{sec:capacity}
We next test whether the failure predicted by the source marginal is driven by transport capacity rather than by the representation or the transport solver. We first characterize its signature, then ask where the displaced mass goes, and finally test which components of the OT formulation can remove it.

\paragraph{Data} We evaluate the capacity hypothesis in two complementary domain-shift settings. In single-cell cross-modal transfer, the source is dissociated RNA sequencing and
the target is spatial transcriptomics from the same tissue, so the domain gap is primarily a change in measurement process \citep{li2022benchmarking}. We use Jint and Janesick \citep{janesick2023xenium}, together with SpatialCorpus \citep{tejada2025nicheformer} as a no-composition-shift control. In image domain adaptation, we use Office-Home \citep{venkateswara2017deep}, where the domain gap is primarily stylistic; because the dataset is class balanced, we independently vary class coverage to separate it from representation separability. Full experimental details are given in App.~\ref{app:data}.

\paragraph{Evaluation} We report macro recall, macro-F1, and micro recall. An intervention counts as a fix only if it improves macro-F1 without reducing micro recall. Results are stratified by capacity and evaluated at the target's natural composition unless stated otherwise. Full metric definitions and protocol are given in App.~\ref{app:data}.

\subsection{Capacity signature}
\label{sec:failure}

Under the default marginal on single-cell data, the largest class reaches recall $0.95$ but precision only $0.19$, while a capacity-limited class reaches recall $0.05$ with precision $1.00$ as can be seen in Table~\ref{tab:precision}. The two errors are asymmetric: the largest class absorbs false positives, while the capacity-limited class is rarely predicted but, when predicted, is usually correct. This is consistent with a class being short of transport mass rather than being confused with another class.

\begin{table}[!htb]
\centering
\caption{Per-class recall / precision on single-cell data (Jint), natural target composition, 3 seeds. The default marginal produces the capacity signature: a capacity-limited class has low recall but high precision, while the largest class has high recall but low precision.}
\label{tab:precision}
\footnotesize
\setlength{\tabcolsep}{4pt}
\begin{tabular}{lccccccc}
\toprule
Source marginal & B\_Plasma & T\_NK & Myeloid & Malignant & Epithelial & Endothelial & Stromal\\
\midrule
source frequencies   & .65/.63 & \textbf{.95/.19} & .30/.53 & .64/.55 & .36/.53 & \textbf{.05/1.00} & .09/.89\\
uniform per class    & .73/.66 & .80/.51 & .43/.69 & .66/.63 & .72/\textbf{.26} & .64/\textbf{.29} & .60/.73\\
composition-matched  & .78/.68 & \textbf{.75/.71} & .61/.62 & .61/.78 & .45/.48 & .46/.72 & \textbf{.69/.72}\\
\bottomrule
\end{tabular}
\end{table}

The pattern is not specific to our implementation or read-out. Three independent entropic-OT implementations produce recall between $0.009$ and $0.020$ for the same capacity limited class on identical embeddings, points, labels, folds, and seeds, whereas a $k$NN vote on the same embedding reaches $0.295$ (Table~\ref{tab:appendix}). Thus, the failure persists across OT solvers even when the underlying representation is fixed.

\subsection{Where the displaced mass goes}
\label{sec:mechanism}

The capacity signature could still reflect poor geometry: perhaps starved classes are poorly represented, and their mass simply flows to nearby classes. We therefore compare two explanations for the failure: limited representation geometry and limited transport capacity. If geometry drives the failure, mass should accumulate in classes that are close in the embedding. If capacity drives the failure, mass should instead accumulate in classes with more available source mass.

We test these two possibilities for each below-capacity class $t$ and candidate absorber $c$ by relating the fraction of $t$'s target mass assigned to $c$ to both the source mass available to $c$ and the embedding distance between $t$ and $c$, while controlling for the other. Because capacity and embedding distance are correlated in the data, we use partial Spearman correlations to measure the association of each with absorbed mass while controlling for the other (Table~\ref{tab:mechanism}).

\begin{table}[!htb]
\centering
\caption{Where displaced mass goes. Partial Spearman correlations of absorbed share
with class capacity and embedding distance, each controlling for the other.}
\label{tab:mechanism}
\footnotesize
\begin{tabular}{lcc}
\toprule
 & single-cell (Jint) & Office-Home\\
\midrule
partial $\rho$(absorbed share, \textbf{capacity} $\mid$ distance) & $\mathbf{+0.566}$ & $\mathbf{+0.329}$\\
partial $\rho$(absorbed share, distance $\mid$ capacity)          & $+0.144$ & $-0.206$\\
$n$ (class, absorber) pairs / units                               & $198\,/\,27$ & $14{,}080\,/\,12$\\
\bottomrule
\end{tabular}
\end{table}

Capacity is positively associated with absorbed mass in both domains: after controlling for distance, classes with more available source mass absorb more of the displaced mass. The association with distance is small and inconsistent across domains. The capacity partial is $+0.566$ in single-cell data and $+0.329$ on Office-Home, compared with distance partials of $+0.144$ and $-0.206$, respectively (Table~\ref{tab:mechanism}). Thus, displaced mass follows available capacity rather than embedding proximity.

This result is consistent with the categorical analysis: in $30$ of $33$ below-capacity observations, the absorbing class is the largest source class, while it is never the nearest class. Repeating the analysis with a PCA-based representation gives the same result (Table~\ref{tab:robust}, App.~\ref{app:additional}). The failure therefore does not disappear when only the representation is improved: the displaced mass is allocated according to the source marginal rather than embedding proximity. App.~\ref{app:visual} shows the same effect in the coupling itself on a single fold.

\subsection{Detection and interventions}
\label{sec:knobs}

The failure is also difficult to detect or repair after the transport plan has been computed: displaced points are often assigned with high confidence to well-supported classes. We therefore test which components of the OT formulation can recover the missing rare-class performance by varying the entropic regularization $\varepsilon$, the marginal penalty $\tau$, and the source marginal $\mu$ under matched conditions. We define a class as below capacity when $r_k<0.15$. We fixed this threshold before the marginal experiments using a preliminary fit of the saturating recall-versus-coverage law of App.~\ref{app:derivation} on held-out single-cell data; the ordering of the main comparisons is unchanged at thresholds of $0.05$, $0.10$, and $0.30$ (Table~\ref{tab:robust}). For the comparison, we use four Office-Home domain pairs and three seeds, yielding $235$ below-capacity and $545$ above-capacity class-observations.

\begin{table}[!htb]
\centering
\caption{Effect of the three OT parameters in Eq.~\ref{eq:uot}, stratified by capacity. Entries are changes in macro recall relative to the default source marginal on Office-Home.}
\label{tab:knobs}
\footnotesize
\begin{tabular}{lcc}
\toprule
intervention & $r_k<0.15$ & $r_k\ge0.15$\\
\midrule
marginal shape (matched to target) & $\mathbf{+0.366}$ & $+0.016$\\
marginal shape (uniform)           & $+0.274$ & $+0.038$\\
$\tau=0.01$ (most relaxed)         & $\mathbf{+0.208}$ & $+0.053$\\
$\tau=\infty$ (balanced Sinkhorn)  & $-0.021$ & $-0.013$\\
$\varepsilon$ ($3\times$ default)  & $\mathbf{+0.000}$ & $-0.012$\\
permuted marginal (control)        & $-0.015$ & $-0.079$\\
\bottomrule
\end{tabular}
\end{table}

The errors are difficult to detect from the transport plan itself. Thirty of $33$ below-capacity observations are confidently wrong, retaining $85\%$ of normal posterior confidence, and five output-side detectors identify only $10$ of $135$ unit evaluations under our fixed detection criterion. Transport entropy is particularly misleading, because displaced points land confidently inside large, well-supported classes. Varying the entropic regularization does not rescue detection either: across seven values, the best detector's selectivity increases from $2.94$ to $23.33$ while below-capacity recall falls from $0.026$ to $0.000$ (Table~\ref{tab:eps}, App.~\ref{app:qa}).

Downstream changes to the transport plan also provide limited recovery. Across ten read-outs of a single converged plan, below-capacity recall increases only from $0.040$ to at most $0.063$ among read-outs based on transported mass (Table~\ref{tab:ceiling}). Across four Office-Home domain pairs and three seeds, changing the entropic regularization produces essentially no gain, while relaxing the marginal penalty provides only partial recovery. Matching the marginal to the target composition is the most effective intervention below capacity, improving macro recall by $+0.366$, compared with $+0.208$ for the most relaxed marginal penalty (Table~\ref{tab:knobs}); above capacity, the effects are small. Relaxing $\tau$ allows the plan to deviate from the source marginal, but it does not determine where the released mass should go, and combining $\tau$ tuning with marginal correction provides no significant additional gain (Table~\ref{tab:mtau}). A permutation control shows that the improvement is not due to changing the marginal arbitrarily: keeping the corrected mass values but assigning them to the wrong classes performs worse than the default marginal on both domains (Table~\ref{tab:knobs}).

These results localize the failure to the class-level mass allocation: the relevant intervention is to assign mass to the classes that need it, rather than to change the read-out or relax the OT regularization.

\section{Correcting the marginal}
\label{sec:correction}

If marginal capacity causes the failure, correcting the source marginal should recover the missing rare-class performance. We first test this with the true target composition, then ask what remains after the correction, and finally evaluate whether the target composition can be estimated well enough for the correction to be used in practice. We use the same datasets and evaluation protocol introduced in Sec.~\ref{sec:capacity}.

\subsection{What the correction repairs}
\label{sec:fix}

To correct the capacity mismatch, we assign source mass to each class in proportion to its target composition rather than its source frequency:

\begin{equation}
\mu_j=\frac{\pi_{y_j}}{n^{\mathrm{s}}_{y_j}},
\qquad
\pi_k=\frac{n^{\mathrm{t}}_k}{n^{\mathrm{t}}}.
\label{eq:cmm}
\end{equation}

We first evaluate this marginal using the oracle target composition $\pi$. This oracle setting isolates the effect of correcting the marginal and provides an upper bound on what a marginal-based correction can achieve.

\begin{table}[!htb]
\centering
\caption{Label-transfer performance under different source marginals and target-composition estimates. Results are shown for single-cell data (Jint) and Office-Home under the natural target composition.}
\label{tab:fix}
\footnotesize
\begin{tabular}{lcccc@{\hskip 16pt}cccc}
\toprule
& \multicolumn{4}{c}{single-cell (Jint)} & \multicolumn{4}{c}{Office-Home}\\
\cmidrule(r){2-5}\cmidrule(l){6-9}
estimate of $\ptgt$ & macro & F1 & micro & below & macro & F1 & micro & \%\\
\midrule
source frequencies & $0.433$ & $0.322$ & $0.393$ & $0.030$ & $0.392$ & $0.359$ & $0.437$ & n/a\\
permuted (control) & n/a & n/a & n/a & n/a & $0.333$ & $0.284$ & $0.360$ & n/a\\
uniform            & $0.654$ & $0.512$ & $0.557$ & $0.533$ & $0.501$ & $0.396$ & $0.448$ & $6\%$\\
$k$NN pre-pass     & $0.648$ & $0.556$ & $0.616$ & $0.463$ & $0.487$ & $0.428$ & $0.518$ & $48\%$\\
JCPOT              & $0.585$ & $0.505$ & $0.595$ & $0.280$ & $0.467$ & $0.434$ & $0.537$ & $\mathbf{59\%}$\\
BBSE               & $0.528$ & $0.439$ & $0.451$ & $0.371$ & $0.490$ & $0.421$ & $0.486$ & $29\%$\\
RLLS               & $0.528$ & $0.439$ & $0.451$ & $0.371$ & $0.493$ & $0.423$ & $0.490$ & $31\%$\\
MLLS               & $0.516$ & $0.417$ & $0.409$ & $\mathbf{0.672}$ & $0.494$ & $0.433$ & $0.510$ & $43\%$\\
\midrule
oracle $\ptgt$     & $0.619$ & $\mathbf{0.623}$ & $\mathbf{0.718}$ & $0.546$ & $0.514$ & $\mathbf{0.515}$ & $\mathbf{0.606}$ & n/a\\
\bottomrule
\end{tabular}
\end{table}

The correction substantially improves label transfer. On single-cell data, macro-F1 increases from $0.322$ to $0.623$, while micro recall increases from $0.393$ to $0.718$. For the single-cell setting, we use three folds and three seeds; for Office-Home, we use four domain pairs and three seeds (Table~\ref{tab:fix}). Uniformly balancing the source mass across classes does not provide the same correction: it assumes a uniform target composition and can improve rare-class recall by sacrificing the dominant class. On one Janesick replicate, uniform balancing increases macro recall while reducing micro recall from $0.477$ to $0.346$, with the largest target class falling from $0.472$ to $0.069$ (Table~\ref{tab:jan}, App.~\ref{app:additional}).

The benefit depends on coverage. It decreases monotonically with the coverage ratio and changes sign above capacity (Fig.~\ref{fig:dose}, App.~\ref{app:additional}). Across Office-Home, the correction ranges from $+0.473$ to $-0.178$ across coverage deciles, while the single-cell data range from $+0.531$ to $-0.355$ across five coverage bins. The crossing point is dataset-dependent, so we do not treat the capacity threshold as a universal break-even point. The same qualitative trend is obtained with a deployable estimate of the target composition, with smaller gains and a later break-even point (Fig.~\ref{fig:estpi}, App.~\ref{app:estpi}).

\subsection{Separating capacity from geometry}
\label{sec:decomp}

Correcting the marginal removes the capacity constraint, so the remaining error should reflect limitations of the representation. We test this by decomposing each class's error into a \emph{removable} component, the gain from the composition-matched marginal, and a \emph{residual} component, the error that remains after correction.

We use Office-Home for this analysis because its class-balanced design lets us vary coverage independently of representation separability. Across $780$ class observations, the removable component is strongly associated with coverage ($\rho=-0.809$) but not with separability ($\rho=-0.009$), whereas the residual component is associated with separability ($\rho=-0.337$) but not with coverage ($\rho=-0.005$) (Fig.~\ref{fig:diss}, App.~\ref{app:additional}).

This double dissociation shows that correcting the marginal removes a coverage-dependent component of transfer error while leaving a residual component associated with class separability. The residual is therefore not simply a consequence of class rarity: some well-covered classes remain poorly transferred when their representation is not separable.

The decomposition is defined using the oracle target composition. Repeating the analysis with estimated target compositions preserves the same qualitative asymmetry across estimators, although the exact dissociation is weaker (Table~\ref{tab:estpi}, App.~\ref{app:estpi}).

\vspace{-1mm}
\subsection{Estimating the target composition}
\label{sec:estimation}

The composition-matched marginal requires the target composition $\pi$, which is unknown in practice. We therefore compare estimators of $\pi$ by feeding each estimate into the same marginal construction, so that differences in performance reflect the composition estimate rather than the transport procedure. We consider three label-shift estimators, BBSE \citep{lipton2018bbse}, RLLS \citep{azizzadenesheli2019rlls}, and MLLS/EM \citep{saerens2002adjusting,alexandari2020maximum}, and two transport-native estimators, JCPOT \citep{redko2019jcpot} and a $k$NN pre-pass.

The label-shift estimators perform similarly and recover little of the oracle gain under the measurement-process shift. On single-cell data, BBSE, RLLS, and MLLS close only $4.7$--$17.8\%$ of the $+0.324$ micro-recall gap to the oracle, compared with $29$--$43\%$ on Office-Home. Their similar performance across three datasets indicates an assumption-level limitation rather than an optimization issue: these methods assume that $p(x\mid y)$ is invariant across domains, an assumption violated by a change in measurement process.

Transport-native estimators perform better on both domains, although the best estimator depends on the shift. The $k$NN pre-pass closes $69\%$ of the oracle gap on single-cell data, while JCPOT closes $59\%$ on Office-Home. No estimator is consistently best across the two domain shifts. We additionally demonstrate that the correction also does not require an exact composition estimate: an estimate only $30\%$ of the way from the source composition toward the target already closes half of the oracle gain (App.~\ref{app:alpha}).

\vspace{-1mm}
\section{Discussion and Conclusion}
\label{sec:discussion}

\vspace{-2mm}
We asked whether the poor transfer of rare classes under composition shift can arise from the OT formulation itself, rather than only from the representation. Our results show that it can. The source marginal imposes a class-level transport capacity, so a class can be starved when its target demand exceeds its available mass, even when the embedding separates it well. The displaced mass is absorbed by classes with greater capacity rather than by geometrically similar classes, and the resulting errors can remain highly confident. Correcting the marginal removes this coverage-dependent component of the error, while the residual tracks class separability. Thus, marginal capacity is a distinct source of transfer error, not simply another manifestation of poor representation quality.

The location of this failure also explains why downstream corrections are limited. Once the coupling has allocated too little mass to a class, changing the read-out, smoothing the plan, or relaxing the marginal constraint cannot reliably recover the missing mass. The effective intervention is instead to change the class-level mass allocation itself. This also clarifies the practical implication of the correction: the target composition, rather than a generic preference for rare classes, should determine the transport marginal.

\vspace{-2mm}
\paragraph{Limitations and Future Work} The composition-matched marginal uses the true target composition and therefore provides an upper bound; estimated compositions weaken the capacity--geometry separation. Future work could develop more accurate target-composition estimators to improve the practical correction. Our analysis treats the representation as fixed; extending OT label transfer to jointly learned, multimodal representations could provide a complementary way to improve cross-domain separability. Finally, the controlled Office-Home experiments provide the clearest evidence for separating coverage from separability, while the single-cell datasets contain fewer below-capacity classes; broader domain shifts and coupling-level diagnostics could test and detect the same failure more generally.

\vspace{-2mm}
\paragraph{Conclusion} OT label transfer under composition shift is limited by two distinct factors: the representation determines which classes can be distinguished, while the transport marginal determines how much mass each class can receive. A class can therefore be under-predicted because the coupling cannot allocate enough mass even when the representation is informative, producing confident errors that downstream read-outs cannot recover. Matching the marginal to the target composition removes this capacity-driven component and reveals the residual geometric error, but the correction is useful primarily where coverage is limiting and requires an estimate of the target composition.

\newpage
\bibliography{neurips_2026}

@inproceedings{lipton2018bbse,
  title     = {Detecting and Correcting for Label Shift with Black Box Predictors},
  author    = {Lipton, Zachary C. and Wang, Yu-Xiang and Smola, Alexander J.},
  booktitle = {Proceedings of the 35th International Conference on Machine Learning (ICML)},
  volume    = {80},
  series    = {PMLR},
  pages     = {3122--3130},
  year      = {2018}
}

@inproceedings{azizzadenesheli2019rlls,
  title     = {Regularized Learning for Domain Adaptation under Label Shifts},
  author    = {Azizzadenesheli, Kamyar and Liu, Anqi and Yang, Fanny and Anandkumar, Animashree},
  booktitle = {International Conference on Learning Representations (ICLR)},
  year      = {2019}
}

@article{saerens2002adjusting,
  title   = {Adjusting the Outputs of a Classifier to New a Priori Probabilities:
             A Simple Procedure},
  author  = {Saerens, Marco and Latinne, Patrice and Decaestecker, Christine},
  journal = {Neural Computation},
  volume  = {14},
  number  = {1},
  pages   = {21--41},
  year    = {2002},
  doi     = {10.1162/089976602753284446}
}

@inproceedings{alexandari2020maximum,
  title     = {Maximum Likelihood with Bias-Corrected Calibration is Hard-to-Beat
               at Label Shift Adaptation},
  author    = {Alexandari, Amr and Kundaje, Anshul and Shrikumar, Avanti},
  booktitle = {Proceedings of the 37th International Conference on Machine Learning (ICML)},
  volume    = {119},
  series    = {PMLR},
  pages     = {222--232},
  year      = {2020}
}

@inproceedings{tachetdescombes2020domain,
  title     = {Domain Adaptation with Conditional Distribution Matching and
               Generalized Label Shift},
  author    = {Tachet des Combes, Remi and Zhao, Han and Wang, Yu-Xiang and Gordon, Geoffrey J.},
  booktitle = {Advances in Neural Information Processing Systems (NeurIPS)},
  year      = {2020}
}

@inproceedings{redko2019jcpot,
  title     = {Optimal Transport for Multi-source Domain Adaptation under Target Shift},
  author    = {Redko, Ievgen and Courty, Nicolas and Flamary, R{\'e}mi and Tuia, Devis},
  booktitle = {Proceedings of the 22nd International Conference on Artificial
               Intelligence and Statistics (AISTATS)},
  volume    = {89},
  series    = {PMLR},
  pages     = {849--858},
  year      = {2019}
}

@inproceedings{cuturi2013sinkhorn,
  title     = {Sinkhorn Distances: Lightspeed Computation of Optimal Transport},
  author    = {Cuturi, Marco},
  booktitle = {Advances in Neural Information Processing Systems (NeurIPS)},
  volume    = {26},
  year      = {2013}
}

@article{peyre2019computational,
  title   = {Computational Optimal Transport: With Applications to Data Science},
  author  = {Peyr{\'e}, Gabriel and Cuturi, Marco},
  journal = {Foundations and Trends in Machine Learning},
  volume  = {11},
  number  = {5--6},
  pages   = {355--607},
  year    = {2019},
  doi     = {10.1561/2200000073}
}

@article{chizat2018scaling,
  title   = {Scaling Algorithms for Unbalanced Optimal Transport Problems},
  author  = {Chizat, Lena{\"i}c and Peyr{\'e}, Gabriel and Schmitzer, Bernhard and
             Vialard, Fran{\c c}ois-Xavier},
  journal = {Mathematics of Computation},
  volume  = {87},
  number  = {314},
  pages   = {2563--2609},
  year    = {2018},
  doi     = {10.1090/mcom/3303}
}

@article{sejourne2019sinkhorn,
  title={Sinkhorn divergences for unbalanced optimal transport},
  author={S{\'e}journ{\'e}, Thibault and Feydy, Jean and Vialard, Fran{\c{c}}ois-Xavier and Trouv{\'e}, Alain and Peyr{\'e}, Gabriel},
  journal={arXiv preprint arXiv:1910.12958},
  year={2019}
}

@article{sejourne2023unbalanced,
  title={Unbalanced optimal transport, from theory to numerics},
  author={S{\'e}journ{\'e}, Thibault and Peyr{\'e}, Gabriel and Vialard, Fran{\c{c}}ois-Xavier},
  journal={Handbook of Numerical Analysis},
  volume={24},
  pages={407--471},
  year={2023},
  publisher={Elsevier}
}

@inproceedings{fatras2021unbalanced,
  title     = {Unbalanced Minibatch Optimal Transport; Applications to Domain Adaptation},
  author    = {Fatras, Kilian and S{\'e}journ{\'e}, Thibault and Courty, Nicolas and Flamary, R{\'e}mi},
  booktitle = {Proceedings of the 38th International Conference on Machine Learning (ICML)},
  volume    = {139},
  series    = {PMLR},
  pages     = {3186--3197},
  year      = {2021}
}

@article{courty2017optimal,
  title   = {Optimal Transport for Domain Adaptation},
  author  = {Courty, Nicolas and Flamary, R{\'e}mi and Tuia, Devis and Rakotomamonjy, Alain},
  journal = {IEEE Transactions on Pattern Analysis and Machine Intelligence},
  volume  = {39},
  number  = {9},
  pages   = {1853--1865},
  year    = {2017},
  doi     = {10.1109/TPAMI.2016.2615921}
}

@article{bunne2024optimaltransport,
  title   = {Optimal Transport for Single-Cell and Spatial Omics},
  author  = {Bunne, Charlotte and Schiebinger, Geoffrey and Krause, Andreas and
             Regev, Aviv and Cuturi, Marco},
  journal = {Nature Reviews Methods Primers},
  volume  = {4},
  pages   = {58},
  year    = {2024},
  doi     = {10.1038/s43586-024-00334-2}
}

@article{klein2025mapping,
  title={Mapping cells through time and space with moscot},
  author={Klein, Dominik and Palla, Giovanni and Lange, Marius and Klein, Michal and Piran, Zoe and Gander, Manuel and Meng-Papaxanthos, Laetitia and Sterr, Michael and Saber, Lama and Jing, Changying and others},
  journal={Nature},
  volume={638},
  number={8052},
  pages={1065--1075},
  year={2025},
  publisher={Nature Publishing Group UK London}
}

@article{cao2022uniport,
  title   = {A Unified Computational Framework for Single-Cell Data Integration
             with Optimal Transport},
  author  = {Cao, Kai and Gong, Qiyu and Hong, Yiguang and Wan, Lin},
  journal = {Nature Communications},
  volume  = {13},
  pages   = {7419},
  year    = {2022},
  doi     = {10.1038/s41467-022-35094-8}
}

@article{cao2022glue,
  title   = {Multi-omics Single-Cell Data Integration and Regulatory Inference
             with Graph-Linked Embedding},
  author  = {Cao, Zhi-Jie and Gao, Ge},
  journal = {Nature Biotechnology},
  volume  = {40},
  pages   = {1458--1466},
  year    = {2022},
  doi     = {10.1038/s41587-022-01284-4}
}

@article{stuart2019seurat,
  title   = {Comprehensive Integration of Single-Cell Data},
  author  = {Stuart, Tim and Butler, Andrew and Hoffman, Paul and Hafemeister, Christoph and
             Papalexi, Efthymia and Mauck, William M. and Hao, Yuhan and Stoeckius, Marlon and
             Smibert, Peter and Satija, Rahul},
  journal = {Cell},
  volume  = {177},
  number  = {7},
  pages   = {1888--1902},
  year    = {2019},
  doi     = {10.1016/j.cell.2019.05.031}
}

@article{xu2021scanvi,
  title   = {Probabilistic Harmonization and Annotation of Single-Cell
             Transcriptomics Data with Deep Generative Models},
  author  = {Xu, Chenling and Lopez, Romain and Mehlman, Edouard and Regier, Jeffrey and
             Jordan, Michael I. and Yosef, Nir},
  journal = {Molecular Systems Biology},
  volume  = {17},
  number  = {1},
  pages   = {e9620},
  year    = {2021},
  doi     = {10.15252/msb.20209620}
}

@article{biancalani2021deep,
  title={Deep learning and alignment of spatially resolved single-cell transcriptomes with Tangram},
  author={Biancalani, Tommaso and Scalia, Gabriele and Buffoni, Lorenzo and Avasthi, Raghav and Lu, Ziqing and Sanger, Aman and Tokcan, Neriman and Vanderburg, Charles R and Segerstolpe, {\AA}sa and Zhang, Meng and others},
  journal={Nature methods},
  volume={18},
  number={11},
  pages={1352--1362},
  year={2021},
  publisher={Nature Publishing Group US New York}
}

@article{brbic2022stellar,
  title   = {Annotation of Spatially Resolved Single-Cell Data with {STELLAR}},
  author  = {Brbi{\'c}, Maria and Cao, Kaidi and Hickey, John W. and Tan, Yuqi and
             Snyder, Michael P. and Nolan, Garry P. and Leskovec, Jure},
  journal = {Nature Methods},
  volume  = {19},
  number  = {11},
  pages   = {1411--1418},
  year    = {2022},
  doi     = {10.1038/s41592-022-01651-8}
}

@article{haviv2025covariance,
  title={The covariance environment defines cellular niches for spatial inference},
  author={Haviv, Doron and Rem{\v{s}}{\'\i}k, J{\'a}n and Gatie, Mohamed and Snopkowski, Catherine and Takizawa, Meril and Pereira, Nathan and Bashkin, John and Jovanovich, Stevan and Nawy, Tal and Chaligne, Ronan and others},
  journal={Nature Biotechnology},
  volume={43},
  number={2},
  pages={269--280},
  year={2025},
  publisher={Nature Publishing Group US New York}
}

@article{lotfollahi2022scarches,
  title   = {Mapping Single-Cell Data to Reference Atlases by Transfer Learning},
  author  = {Lotfollahi, Mohammad and Naghipourfar, Mohsen and Luecken, Malte D. and
             Khajavi, Matin and B{\"u}ttner, Maren and Wagenstetter, Marco and others
             and Theis, Fabian J.},
  journal = {Nature Biotechnology},
  volume  = {40},
  pages   = {121--130},
  year    = {2022},
  doi     = {10.1038/s41587-021-01001-7}
}

@article{dominguezconde2022celltypist,
  title   = {Cross-Tissue Immune Cell Analysis Reveals Tissue-Specific Features in Humans},
  author  = {Dom{\'i}nguez Conde, C. and Xu, C. and Jarvis, L. B. and Rainbow, D. B. and
             Wells, S. B. and Gomes, T. and others and Teichmann, S. A.},
  journal = {Science},
  volume  = {376},
  number  = {6594},
  pages   = {eabl5197},
  year    = {2022},
  doi     = {10.1126/science.abl5197}
}

@article{li2022benchmarking,
  title={Benchmarking spatial and single-cell transcriptomics integration methods for transcript distribution prediction and cell type deconvolution},
  author={Li, Bin and Zhang, Wen and Guo, Chuang and Xu, Hao and Li, Longfei and Fang, Minghao and Hu, Yinlei and Zhang, Xinye and Yao, Xinfeng and Tang, Meifang and others},
  journal={Nature methods},
  volume={19},
  number={6},
  pages={662--670},
  year={2022},
  publisher={Nature Publishing Group US New York}
}

@article{janesick2023xenium,
  title   = {High Resolution Mapping of the Tumor Microenvironment Using Integrated
             Single-Cell, Spatial and In Situ Analysis},
  author  = {Janesick, Amanda and Shelansky, Robert and Gottscho, Andrew D. and
             Wagner, Florian and Williams, Stephen R. and Rouault, Morgane and others},
  journal = {Nature Communications},
  volume  = {14},
  pages   = {8353},
  year    = {2023},
  doi     = {10.1038/s41467-023-43458-x}
}

@article{tejada2025nicheformer,
  title={Nicheformer: a foundation model for single-cell and spatial omics},
  author={Tejada-Lapuerta, Alejandro and Schaar, Anna C and Gutgesell, Robert and Palla, Giovanni and Halle, Lennard and Minaeva, Mariia and Vornholz, Larsen and Dony, Leander and Drummer, Francesca and Richter, Till and others},
  journal={Nature methods},
  volume={22},
  number={12},
  pages={2525--2538},
  year={2025},
  publisher={Nature Publishing Group US New York}
}

@article{cui2024scgpt,
  title   = {{scGPT}: Toward Building a Foundation Model for Single-Cell Multi-omics
             Using Generative AI},
  author  = {Cui, Haotian and Wang, Chloe and Maan, Hassaan and Pang, Kuan and
             Luo, Fengning and Duan, Nan and Wang, Bo},
  journal = {Nature Methods},
  volume  = {21},
  pages   = {1470--1480},
  year    = {2024},
  doi     = {10.1038/s41592-024-02201-0}
}

@article{hao2024scfoundation,
  title   = {Large-Scale Foundation Model on Single-Cell Transcriptomics},
  author  = {Hao, Minsheng and Gong, Jing and Zeng, Xin and Liu, Chiming and Guo, Yucheng and
             Cheng, Xingyi and Wang, Taifeng and Ma, Jianzhu and Zhang, Xuegong and Song, Le},
  journal = {Nature Methods},
  volume  = {21},
  pages   = {1481--1491},
  year    = {2024},
  doi     = {10.1038/s41592-024-02305-7}
}

@article{theodoris2023geneformer,
  title   = {Transfer Learning Enables Predictions in Network Biology},
  author  = {Theodoris, Christina V. and Xiao, Ling and Chopra, Anant and Chaffin, Mark D. and
             Al Sayed, Zeina R. and Hill, Matthew C. and Mantineo, Helene and others and
             Ellinor, Patrick T.},
  journal = {Nature},
  volume  = {618},
  pages   = {616--624},
  year    = {2023},
  doi     = {10.1038/s41586-023-06139-9}
}

@article{rosen2023uce,
  title   = {Universal Cell Embeddings: A Foundation Model for Cell Biology},
  author  = {Rosen, Yanay and Roohani, Yusuf and Agrawal, Ayush and Samotorcan, Leon and
             {Tabula Sapiens Consortium} and Quake, Stephen R. and Leskovec, Jure},
  journal = {bioRxiv},
  year    = {2023},
  doi     = {10.1101/2023.11.28.568918}
}

@inproceedings{wen2024cellplm,
  title     = {{CellPLM}: Pre-training of Cell Language Model Beyond Single Cells},
  author    = {Wen, Hongzhi and Tang, Wenzhuo and Dai, Xinnan and Ding, Jiayuan and
               Jin, Wei and Xie, Yuying and Tang, Jiliang},
  booktitle = {International Conference on Learning Representations (ICLR)},
  year      = {2024}
}

@article{kedzierska2025zeroshot,
  title   = {Zero-Shot Evaluation Reveals Limitations of Single-Cell Foundation Models},
  author  = {Kedzierska, Kasia Z. and Crawford, Lorin and Amini, Ava P. and Lu, Alex X.},
  journal = {Genome Biology},
  volume  = {26},
  pages   = {101},
  year    = {2025},
  doi     = {10.1186/s13059-025-03574-x}
}

@article{ahlmanneltze2025linear,
  title   = {Deep-Learning-Based Gene Perturbation Effect Prediction Does Not Yet
             Outperform Simple Linear Baselines},
  author  = {Ahlmann-Eltze, Constantin and Huber, Wolfgang and Anders, Simon},
  journal = {Nature Methods},
  volume  = {22},
  pages   = {1657--1661},
  year    = {2025},
  doi     = {10.1038/s41592-025-02772-6}
}

@article{boiarsky2023deepdive,
  title   = {A Deep Dive into Single-Cell {RNA} Sequencing Foundation Models},
  author  = {Boiarsky, Rebecca and Singh, Nalini and Buendia, Alejandro and
             Getz, Gad and Sontag, David},
  journal = {bioRxiv},
  year    = {2023},
  doi     = {10.1101/2023.10.19.563100}
}

@article{bendidi2024pca,
  title   = {Benchmarking Transcriptomics Foundation Models for Perturbation Analysis:
             One {PCA} Still Rules Them All},
  author  = {Bendidi, Ihab and Whitfield, Shawn and Kenyon-Dean, Kian and
             Ben Yedder, Hanene and El Mesbahi, Yassir and Noutahi, Emmanuel and
             Denton, Alisandra Kaye},
  journal = {arXiv preprint arXiv:2410.13956},
  year    = {2024},
  note    = {NeurIPS 2024 AIDrugX Workshop}
}

@article{wenteler2024perteval,
  title={PertEval-scFM: benchmarking single-cell foundation models for perturbation effect prediction},
  author={Wenteler, Aaron and Occhetta, Martina and Branson, Nikhil and Huebner, Magdalena and Curean, Victor and Dee, WT and Connell, WT and Hawkins-Hooker, Alex and Chung, Siu Pui and Ektefaie, Yasha and others},
  journal={BioRxiv},
  pages={2024--10},
  year={2024},
  publisher={Cold Spring Harbor Laboratory}
}

@inproceedings{guo2017calibration,
  title     = {On Calibration of Modern Neural Networks},
  author    = {Guo, Chuan and Pleiss, Geoff and Sun, Yu and Weinberger, Kilian Q.},
  booktitle = {Proceedings of the 34th International Conference on Machine Learning (ICML)},
  volume    = {70},
  series    = {PMLR},
  pages     = {1321--1330},
  year      = {2017}
}

@inproceedings{nguyen2015deep,
  title     = {Deep Neural Networks Are Easily Fooled: High Confidence Predictions
               for Unrecognizable Images},
  author    = {Nguyen, Anh and Yosinski, Jason and Clune, Jeff},
  booktitle = {IEEE Conference on Computer Vision and Pattern Recognition (CVPR)},
  pages     = {427--436},
  year      = {2015}
}

@inproceedings{ovadia2019trust,
  title     = {Can You Trust Your Model's Uncertainty? Evaluating Predictive
               Uncertainty Under Dataset Shift},
  author    = {Ovadia, Yaniv and Fertig, Emily and Ren, Jie and Nado, Zachary and
               Sculley, D. and Nowozin, Sebastian and Dillon, Joshua V. and
               Lakshminarayanan, Balaji and Snoek, Jasper},
  booktitle = {Advances in Neural Information Processing Systems (NeurIPS)},
  year      = {2019}
}

@inproceedings{geifman2017selective,
  title     = {Selective Classification for Deep Neural Networks},
  author    = {Geifman, Yonatan and El-Yaniv, Ran},
  booktitle = {Advances in Neural Information Processing Systems (NeurIPS)},
  year      = {2017}
}

@inproceedings{hendrycks2017baseline,
  title     = {A Baseline for Detecting Misclassified and Out-of-Distribution
               Examples in Neural Networks},
  author    = {Hendrycks, Dan and Gimpel, Kevin},
  booktitle = {International Conference on Learning Representations (ICLR)},
  year      = {2017}
}

@book{vovk2005algorithmic,
  title     = {Algorithmic Learning in a Random World},
  author    = {Vovk, Vladimir and Gammerman, Alexander and Shafer, Glenn},
  publisher = {Springer},
  year      = {2005}
}

@article{angelopoulos2021gentle,
  title   = {A Gentle Introduction to Conformal Prediction and Distribution-Free
             Uncertainty Quantification},
  author  = {Angelopoulos, Anastasios N. and Bates, Stephen},
  journal = {arXiv preprint arXiv:2107.07511},
  year    = {2021}
}

@article{srinivasan2022michaelis,
  title   = {A Guide to the Michaelis--Menten Equation: Steady State and Beyond},
  author  = {Srinivasan, Bharath},
  journal = {The FEBS Journal},
  volume  = {289},
  number  = {20},
  pages   = {6086--6098},
  year    = {2022},
  doi     = {10.1111/febs.16124}
}

@article{hestness2017deep,
  title   = {Deep Learning Scaling Is Predictable, Empirically},
  author  = {Hestness, Joel and Narang, Sharan and Ardalani, Newsha and Diamos, Gregory and
             Jun, Heewoo and Kianinejad, Hassan and Patwary, Md. Mostofa Ali and
             Yang, Yang and Zhou, Yanqi},
  journal = {arXiv preprint arXiv:1712.00409},
  year    = {2017}
}

@article{kaplan2020scaling,
  title   = {Scaling Laws for Neural Language Models},
  author  = {Kaplan, Jared and McCandlish, Sam and Henighan, Tom and Brown, Tom B. and
             Chess, Benjamin and Child, Rewon and Gray, Scott and Radford, Alec and
             Wu, Jeffrey and Amodei, Dario},
  journal = {arXiv preprint arXiv:2001.08361},
  year    = {2020}
}

@article{viering2023shape,
  title   = {The Shape of Learning Curves: A Review},
  author  = {Viering, Tom and Loog, Marco},
  journal = {IEEE Transactions on Pattern Analysis and Machine Intelligence},
  volume  = {45},
  number  = {6},
  pages   = {7799--7819},
  year    = {2023},
  doi     = {10.1109/TPAMI.2022.3220744}
}

@article{chawla2002smote,
  title   = {{SMOTE}: Synthetic Minority Over-sampling Technique},
  author  = {Chawla, Nitesh V. and Bowyer, Kevin W. and Hall, Lawrence O. and
             Kegelmeyer, W. Philip},
  journal = {Journal of Artificial Intelligence Research},
  volume  = {16},
  pages   = {321--357},
  year    = {2002}
}

@article{dablain2022deepsmote,
  title={DeepSMOTE: Fusing deep learning and SMOTE for imbalanced data},
  author={Dablain, Damien and Krawczyk, Bartosz and Chawla, Nitesh V},
  journal={IEEE transactions on neural networks and learning systems},
  volume={34},
  number={9},
  pages={6390--6404},
  year={2022},
  publisher={IEEE}
}

@inproceedings{el2021label,
  title={Label propagation through optimal transport},
  author={El Hamri, Mourad and Bennani, Youn{\`e}s and Falih, Issam},
  booktitle={2021 International Joint Conference on Neural Networks (IJCNN)},
  pages={1--8},
  year={2021},
  organization={IEEE}
}

@inproceedings{venkateswara2017deep,
  title     = {Deep Hashing Network for Unsupervised Domain Adaptation},
  author    = {Venkateswara, Hemanth and Eusebio, Jose and Chakraborty, Shayok and
               Panchanathan, Sethuraman},
  booktitle = {IEEE Conference on Computer Vision and Pattern Recognition (CVPR)},
  pages     = {5018--5027},
  year      = {2017}
}
\bibliographystyle{plainnat}

\newpage
\appendix

\section*{Reproducibility statement}
Apps.~\ref{app:derivation} and~\ref{app:data} give the derivation with its
assumptions and scope, and the datasets, folds, budgets, solver settings and
metric definitions. The analysis in App.~\ref{app:estpi} re-reads logged per-class recalls from the runs behind Table~\ref{tab:fix} and solves no new transport problem. Janesick, SpatialCorpus and Office-Home are public, and every result in this paper can be reproduced from the specifications given here: Sec.~\ref{sec:capacity} and App.~\ref{app:data} fix the encoder, cost, marginals, solver and convergence criterion, the folds, seeds and source budgets, the target-composition design of each arm, and the metric definitions, and Apps.~\ref{app:qa} to~\ref{app:additional} report every arm we ran rather than a selected subset.

\section{Extended related work}
\label{app:related}

\paragraph{Transport-based label transfer.} Unpaired modalities are aligned with
coupled autoencoders and OT \citep{cao2022uniport}, graph-linked embeddings
\citep{cao2022glue}, anchor transfer \citep{stuart2019seurat}, semi-supervised
generative models \citep{xu2021scanvi}, geometric deep learning
\citep{brbic2022stellar} and niche-covariance models \citep{haviv2025covariance};
reference mapping transfers atlas labels
\citep{lotfollahi2022scarches,dominguezconde2022celltypist}; foundation models
supply reusable embeddings
\citep{tejada2025nicheformer,cui2024scgpt,hao2024scfoundation,theodoris2023geneformer,rosen2023uce,wen2024cellplm}.
None characterizes, per class, when transfer is feasible, and none identifies the
source marginal as the controlling object.

\paragraph{Entropic and unbalanced OT.} Entropic OT
\citep{cuturi2013sinkhorn,peyre2019computational} and its unbalanced relaxations
\citep{chizat2018scaling,sejourne2019sinkhorn,sejourne2023unbalanced,fatras2021unbalanced}
underpin single-cell mapping
\citep{biancalani2021deep,klein2025mapping,bunne2024optimaltransport} and domain
adaptation \citep{courty2017optimal}. The unbalanced relaxation is motivated
throughout as the response to composition mismatch. Sec.~\ref{sec:knobs} measures
that response against specifying the marginal correctly and finds the latter
roughly twice as effective on the classes that have a deficit.

\paragraph{Label shift.} See Sec.~\ref{sec:estimation} for positioning.
Generalized label shift \citep{tachetdescombes2020domain} relaxes the invariance
assumption that Sec.~\ref{sec:estimation} finds violated under a
measurement-process gap. Abundance-aware spot deconvolution
\citep{li2022benchmarking} is the nearest acknowledgement in spatial
transcriptomics that composition is first-class, but it estimates mixture
proportions per location rather than labels per point.

\paragraph{Simple baselines, confident errors, saturation laws.} Foundation-model
embeddings frequently fail to beat simple baselines
\citep{kedzierska2025zeroshot,boiarsky2023deepdive,bendidi2024pca,ahlmanneltze2025linear,wenteler2024perteval};
Sec.~\ref{sec:correction} shows one such reported gap dissolving under matched
corrections. Classifiers are miscalibrated off-distribution
\citep{guo2017calibration,nguyen2015deep,ovadia2019trust}, and selective
prediction \citep{geifman2017selective}, softmax and OOD baselines
\citep{hendrycks2017baseline} and conformal prediction
\citep{vovk2005algorithmic,angelopoulos2021gentle} act at the output, where
Sec.~\ref{sec:mechanism} finds these errors invisible. Equation~\ref{eq:law} is
a rectangular hyperbola of Michaelis and Menten form
\citep{srinivasan2022michaelis}, derived from the marginal rather than fitted as
an empirical scaling law
\citep{hestness2017deep,kaplan2020scaling,viering2023shape}. Optimal-Transport
Propagation, which we test in App.~\ref{app:qa}, diffuses labels over a graph
built from the coupling and is the strongest plan-side read-out we are aware of.

\section{The saturating law and its scope}
\label{app:derivation}

For a target point $x$ of true class $t$, the entropic row of the optimal plan is
$T_{x\cdot}\propto\exp(-(C_{x\cdot}-f_x-g)/\varepsilon)$. Aggregating by source
label, $\sum_{j\in t}T_{xj}\propto\nsrc_t\,\bar{e}_t$ with
$\bar{e}_t\approx e^{-\sigma^{2}/\varepsilon}$ for the true class and
$\bar{e}_c\approx e^{-d^{2}/\varepsilon}$ for the nearest well-covered class $c$,
where $\sigma$ is within-class scatter and $d$ centroid separation. A soft
read-out assigns $t$ with probability
$\nsrc_t\bar{e}_t/(\nsrc_t\bar{e}_t+\nsrc_c\bar{e}_c)$; dividing by $\ntgt_t$
gives
\begin{equation}
\mathrm{recall}(t)\;\approx\;\frac{a\,r_t}{\rstar+r_t},\qquad
\rstar=\kappa\,e^{-(d^{2}-\sigma^{2})/\varepsilon}.
\label{eq:law}
\end{equation}
Note the two factors: a capacity term in $r_t$ and a geometry term in
$d^2-\sigma^2$. The double dissociation of Sec.~\ref{sec:decomp} is a test of exactly
this structure, and it is the part of Eq.~\ref{eq:law} that survives outside the
single-cell setting.

\paragraph{Assumptions and scope.} The derivation is written for a \emph{soft} read-out while the deployed one is an $\arg\max$, and in the $\tau\to\infty$ limit, absorbing the marginal penalty into the potentials; with finite $\tau$ mass conservation is soft, so Eq.~\ref{eq:law} is the hard-constraint statement. The $\tau$ sweep is consistent with this, since the pathology is strongest exactly at $\tau\to\infty$. The form does not transfer to Office-Home, where the median $R^2$ is $0.276$ for transport and $0.142$ for the vote, $\rstar$ is undefined for roughly half the classes under transport, the transport-to-vote ratio is $1.26\times$, and no class sits at chance (Table~\ref{tab:lawfits}). We therefore present it as a single-cell instantiation rather than a general law. We do not quote a single-cell threshold ratio: the logged Janesick fits predate the $a\le1$ convention, no post-refit value exists in a logged run, and at the default $\tau$ the fits are not identified at all (Sec.~\ref{sec:knobs}, App.~\ref{app:qa}). The single-cell claim therefore rests on the fit-free resolvability count of Sec.~\ref{sec:failure}, which is computed before any fit is attempted. What fails on Office-Home is the parametric \emph{form}. The structure it is derived from, a capacity factor in $r$ times a geometry factor, is tested directly in Sec.~\ref{sec:decomp} and holds on both domains. The qualitative gap has the same explanation as the chance-level classes of Sec.~\ref{sec:failure}: a $65$-class tail has no dominant absorber, so no class is starved hard enough to sit at chance.

\begin{table}[ht]
\centering
\caption{The saturating law fitted on Office-Home, transport read-out against a
$k$NN vote. Same protocol as the single-cell fits: per class,
$\mathrm{recall}(t)\approx a\,r_t/(\rstar+r_t)$, ceiling constrained to $a\le1$,
and any fit landing within $10^{-3}$ of that bound declined. No fit is pinned here
(the largest ceiling is $0.9951$), so nothing is declined for that reason. The
form does not transfer: the fits are poor, a threshold is undefined for roughly
half the classes under transport, the transport-to-vote ratio is close to unity
with a range spanning both sides of it, and no class sits at chance. Single-cell
fits are not tabulated alongside, because the logged Janesick fits predate the
$a\le1$ convention; Sec.~\ref{sec:failure} rests the single-cell claim on the
fit-free resolvability count instead, which no fitting convention can affect.}
\label{tab:lawfits}
\footnotesize
\begin{tabular}{lcc}
\toprule
 & transport & $k$NN vote\\
\midrule
classes resolvable                         & $65/65$ & $64/65$\\
classes with a finite $\rstar$             & $36/65$ & $47/65$\\
median $R^2$                               & $0.276$ & $0.142$\\
median $\rstar$                            & $0.142$ & $0.108$\\
median ceiling $a$                         & $0.657$ & $0.606$\\
fits declined at the $a\le1$ bound         & $0/36$  & $0/47$\\
classes at chance at every coverage tested & $0/65$  & $0/65$\\
\midrule
\multicolumn{3}{l}{\emph{paired over the 34 classes resolving under both}}\\
median $\rstar$ ratio, transport / vote    & \multicolumn{2}{c}{$1.263\times$}\\
range of the ratio                         & \multicolumn{2}{c}{$[0.225\times,\;3.560\times]$}\\
\bottomrule
\end{tabular}
\end{table}

\section{Data, protocol and metrics}
\label{app:data}

\paragraph{Jint (in-house).} Twelve paired spatial and dissociated slides;
$802{,}132$ dissociated points $\times$ $18{,}051$ features and
$27{,}238$ to $273{,}070$ points per spatial slide $\times$ $5{,}001$ features;
seven reconciled classes; annotations on both sides with zero missing labels. Main
runs use folds carrying a below-capacity class, an uncapped $195{,}598$-point
source, three seeds, and a converged Sinkhorn--Knopp and
majorization--minimization chain, the two agreeing in objective to eight decimals
at $5\times10^{5}$ iterations, with every run recording its converged backend. The
below-capacity stratum is three (fold, class) cells.

\paragraph{Janesick (public, GSE243275).} Pooled $3'$ and $5'$ dissociated source
($22{,}969$ and $12{,}388$ points $\times$ $36{,}601$ features) to a
$313$-feature spatial target ($167{,}780$ and $118{,}752$ points, two replicates);
published annotations on all four files. Per-class law fits use $48$ observations;
the coverage gradient spans three orders of magnitude and includes an
above-capacity control class ($r=0.601$).

\paragraph{SpatialCorpus (public atlas).} Five tissues, two species; source and
target are partitions of one dissociated file, so compositions agree by
construction. Used as the composition-shift negative control.

\paragraph{Office-Home.} Four domains, $65$ classes, $15{,}588$ images; CLIP
ViT-B/32 features, with ResNet-50 followed by PCA as a second backbone. Target
composition long-tailed at $100{:}1$; per-class $r$ drawn log-uniform on
$[0.01,2]$ and $\nsrc_t$ derived from the draw, so the realised ratio matches the
design to $2\times10^{-16}$, where dividing two independent long tails instead
gives $[0.014,39]$. Achieved $r\in[0.020,5.50]$ with $235/780$ class-observations
below capacity, and $\rho(\log r,\text{separability})=+0.108$. Four domain pairs,
three seeds. Each Office-Home experiment draws its own composition profile, held
fixed across all arms within that experiment, so every comparison we report is a
within-experiment contrast and we do not compare arms across experiments.
$\varepsilon$ was recalibrated on CLIP features rather than carried over.

\paragraph{Source budgets.} Training-free read-outs use the uncapped pool. Trained
methods cannot fit it and use a capped proportional source; composition is
preserved while absolute counts scale, so the two are reported separately and
never ranked together. Backbone-invariance runs use a $40{,}000$-point source
because the PCA basis needs counts, and they test reproduction on another
representation rather than an absolute $\rstar$.

\paragraph{Target-composition design, by result.} Natural composition is used for
the marginal and estimator panels (Table~\ref{tab:fix}), the matched-correction
comparison (Table~\ref{tab:fair}), every Office-Home result, and the mechanism
counts and capacity partials (Table~\ref{tab:mechanism}). A fixed-per-class target
is used, and enters only as a within-run contrast, for the four-solver comparison
and the detector suite (Table~\ref{tab:appendix}), the backbone-invariance run,
the $\varepsilon$ and $\tau$ sweeps and the permutation control on single-cell
data (Sec.~\ref{sec:knobs}), the read-out and detection experiments of
App.~\ref{app:qa}, the ratio-versus-count pair, the minority-synthesis nulls, and
the Janesick per-class curve fits, where the coverage cap is part of the sweep
design. In each of the latter the target is held fixed across the arms compared,
so the ordering is unaffected, while absolute levels are not comparable across the
two designs.

\paragraph{Metrics.} Macro recall is the unweighted mean of per-class recall,
macro-F1 the unweighted mean of per-class F1, and micro recall overall per-point
accuracy. Absorption is the fraction of a below-capacity class's target mass
received by a single other class. Detector selectivity is the ratio of
below-capacity error rate among flagged points to that among unflagged points, and
coverage is the fraction of points not abstained on. Domain identity is close to
linearly separable in the embedding, so we remove it by per-domain mean centering
before computing $C$; this is standard and is itself worth a large fraction of
macro recall ($0.320$ to $0.426$ from raw to centered, absorption $0.994$ to
$0.974$). Every arm in every comparison shares it.

\section{The correction, seen in the plan}
\label{app:visual}

The results in the body are aggregate statistics. This appendix shows the same
effect on a single fold, in the coupling itself and in the embedding, with the
source marginal as the only thing that differs between arms. Both figures use
Jint fold G1, seed $42$, a $195{,}893$-point source and a $5{,}000$-point target
at natural composition, with identical $\varepsilon$ and $\tau=1$ throughout.
Endothelial ($r=0.141$) is the one class below capacity on this fold; Stromal
($r=0.32$) improves by a similar amount but is not below capacity, and we do not
describe it as such.

\begin{figure}[ht]
\centering
\includegraphics[width=\textwidth]{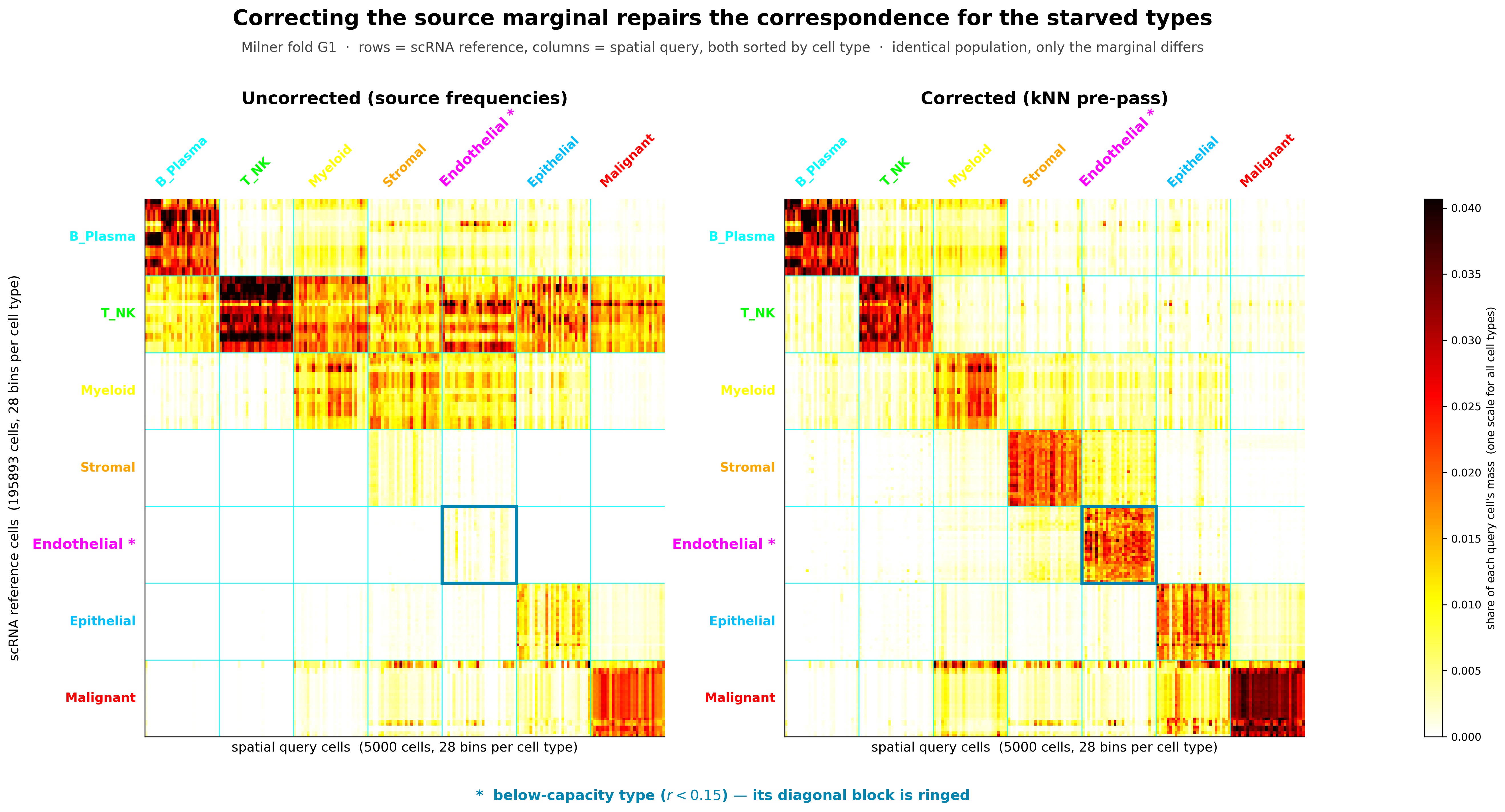}
\caption{\textbf{The failure is in the coupling, and the marginal repairs it.}
Entropic unbalanced OT couplings between the source (rows) and the target
(columns), both sorted by class. Each class is allocated the same number of bins
on each axis, so a class holding $0.3\%$ of the source is as legible as one
holding half of it, and colour is the share of each target point's transported
mass falling in a given source bin, on one scale shared by both panels and all
classes. \emph{Left:} the source marginal is the source's own class frequencies.
\emph{Right:} the marginal is rebuilt as $\mu_j=\hat{\ptgt}_{y_j}/\nsrc_{y_j}$
from a $k$NN pre-pass estimate of the target composition, which uses no target
labels. Everything else is identical. The horizontal band on the T\_NK rows in
the left panel is the absorption mechanism of Sec.~\ref{sec:mechanism} made
visible: target points of a starved class are transported to a class that has
capacity. Endothelial, the one class below capacity here, is starred and its
diagonal block is ringed; uncorrected it sends $46.9\%$ of its mass to T\_NK and
$3.9\%$ to itself, and corrected it sends $51.3\%$ to itself. Aggregated over the
whole target, the T\_NK source rows absorb $47.8\%$ of all transported mass before
correction and $14.4\%$ after.}
\label{fig:coupling}
\end{figure}

\begin{figure}[ht]
\centering
\includegraphics[width=\textwidth]{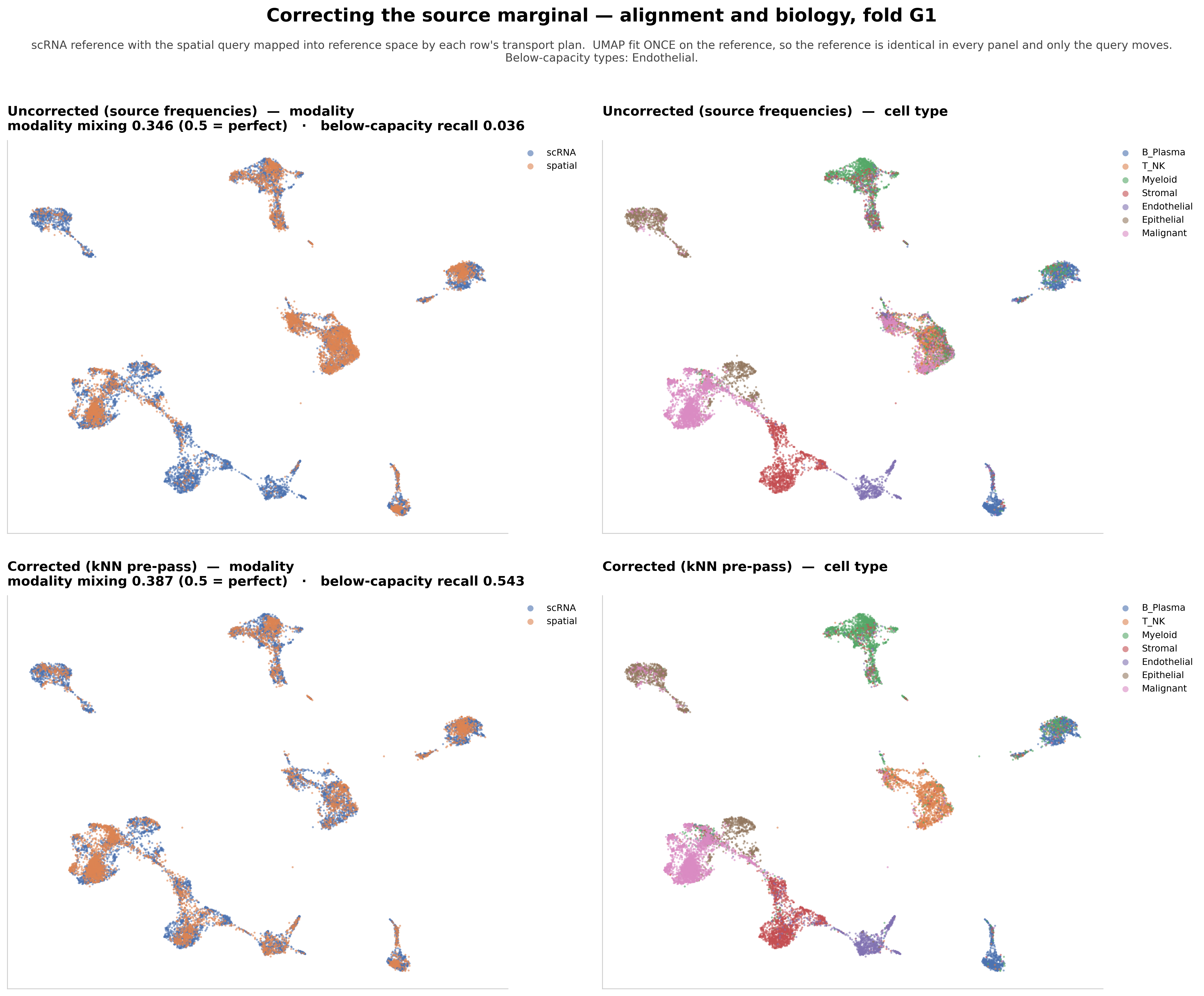}
\caption{\textbf{The correction acts on class correspondence, not on domain
alignment.} UMAP of the source together with the target mapped into source space
by each row's transport plan through barycentric projection. The reducer is fit
once, on the source alone, and both rows are transformed into that fixed
embedding, so the source points are identical in every panel and only the target
moves; any difference between rows is therefore attributable to the plan and not
to a refit. \emph{Top row:} source-frequency marginal. \emph{Bottom row:}
composition-matched marginal from the $k$NN pre-pass. \emph{Left column:} points
coloured by domain, which measures alignment. \emph{Right column:} points
coloured by class, which checks that the biology is preserved. Domain mixing,
the fraction of each point's fifteen nearest neighbours drawn from the other
domain on the plotted coordinates, where $0.5$ is perfect interleaving, improves
only from $0.346$ to $0.387$: the two domains were already reasonably
interleaved, and alignment is not what the marginal repairs. The class panels
show what does change, with the Endothelial population diffuse and largely
mislabelled before correction and consolidated after, and below-capacity recall
rising from $0.036$ to $0.543$.}
\label{fig:umap}
\end{figure}

Figure~\ref{fig:umap} shows the same fold in the embedding rather than the plan.

\paragraph{Per-class detail, and what the correction costs.}
Table~\ref{tab:g1} gives recall and diagonal coupling mass for every class on this
fold. Two things are worth stating plainly. The gain is largest on the classes
with the least coverage and shrinks monotonically as $r$ rises, which is the
dose-response of Sec.~\ref{sec:fix} on a single fold. And the correction is not
free: T\_NK, the class doing the absorbing, falls from $0.948$ to $0.768$, since
the marginal redistributes mass rather than creating it. That is why the
adjudicating rule is macro-F1 with a micro floor rather than below-capacity recall
alone.

\begin{table}[ht]
\centering
\caption{Jint fold G1, per class, ordered by coverage ratio. \emph{Diagonal
mass} is the share of a class's target mass landing on its own source block, read
off the plotted matrices of Fig.~\ref{fig:coupling}. Endothelial is the only class
below capacity on this fold.}
\label{tab:g1}
\footnotesize
\begin{tabular}{lcccc}
\toprule
class & $r$ & recall before & recall after & diagonal mass\\
\midrule
Endothelial (below capacity) & $\mathbf{0.141}$ & $0.036$ & $\mathbf{0.543}$ & $3.9\% \to 51.3\%$\\
Stromal      & $0.320$  & $0.094$ & $0.619$ & $8.7\% \to 55.8\%$\\
Epithelial   & $1.903$  & $0.238$ & $0.514$ & $23.7\% \to 48.6\%$\\
Malignant    & $2.305$  & $0.561$ & $0.896$ & $53.1\% \to 84.9\%$\\
Myeloid      & $3.135$  & $0.314$ & $0.442$ & $32.5\% \to 43.3\%$\\
B\_Plasma    & $5.943$  & $0.711$ & $0.853$ & $69.9\% \to 83.9\%$\\
T\_NK        & $19.875$ & $0.948$ & $\mathbf{0.768}$ & $93.5\% \to 73.4\%$\\
\midrule
macro / micro & & $0.415$ / $0.443$ & $0.662$ / $0.690$ & \\
domain mixing & & $0.346$ & $0.387$ & \\
\bottomrule
\end{tabular}
\end{table}

\section{Read-out, marginal penalty and entropic regularizer, in full}
\label{app:qa}

This appendix reports in full the three tunings summarized in
Secs.~\ref{sec:failure} to \ref{sec:knobs}. Each was specified before it was run,
and each has the same shape: the knob moves the deficit part of the way and none
closes it.

\paragraph{Read-outs of a fixed coupling.} One converged plan per unit, with every
read-out applied to that identical plan, so no two arms can differ by a solver
iteration. The $k$NN vote is the deliberate exception, since it ignores the plan
entirely and is the no-coupling reference. Table~\ref{tab:readout} gives
below-capacity recall for all eleven arms on three datasets.

\begin{table}[ht]
\centering
\caption{Read-out battery, below-capacity recall under the natural marginal. Every
arm except $k$NN consumes the same converged plan. Hard, tempered and soft
read-outs are identical to four decimals; propagation buys a little and twenty
steps buy nothing over one; the only plan read-outs that help are the two that
divide by source class count, which is the marginal correction relocated after the
solve. $n=12$ (Jint), $2{,}412$ (rep1), $1{,}926$ (rep2).}
\label{tab:readout}
\footnotesize
\begin{tabular}{llccc}
\toprule
family & read-out & Jint & Janesick rep1 & Janesick rep2\\
\midrule
hard      & argmax                & $0.040$ & $0.071$ & $0.211$\\
tempered  & softmax $T{=}1$       & $0.040$ & $0.071$ & $0.211$\\
tempered  & softmax $T{=}0.05$    & $0.040$ & $0.071$ & $0.211$\\
soft      & soft posterior        & $0.040$ & $0.071$ & $0.211$\\
\midrule
propagating & OTP, 1 step         & $0.062$ & $0.085$ & $0.254$\\
propagating & OTP, 20 steps       & $0.063$ & $0.086$ & $0.259$\\
propagating & OTP no gate, 1 step & $0.062$ & $0.085$ & $0.254$\\
propagating & OTP no gate, 20 steps & $0.063$ & $0.086$ & $0.259$\\
\midrule
prior-corrected & argmax $/\,\nsrc_t$      & $\mathbf{0.199}$ & $\mathbf{0.180}$ & $0.383$\\
prior-corrected & OTP$_{20}$ $/\,\nsrc_t$  & $0.188$ & $0.190$ & $\mathbf{0.394}$\\
\midrule
plan-free & $k$NN vote            & $\mathbf{0.395}$ & $0.165$ & $\mathbf{0.465}$\\
\bottomrule
\end{tabular}
\end{table}

Two conclusions follow, and a third that we pre-registered and must withdraw.
Softening changes nothing: a soft posterior over a plan whose mass sits on the
wrong classes is a soft posterior over the wrong classes. And a read-out that
never touches the plan beats every read-out that does, on two of three datasets.
The withdrawn prediction concerns the entropy gate inside OTP, which we expected
to suppress rare classes, since confident hallucination implies it should clamp
confident but wrong points early. Gated and ungated variants agree to four
decimals everywhere in Table~\ref{tab:readout}, so the gate contributes nothing
and we report it as a null.

\paragraph{The floor is in the coupling, not in its interpretation.}
Table~\ref{tab:ceiling} settles the question without appeal to any baseline. The
share of transported mass reaching a below-capacity class is under two percent on
every dataset, and the plain argmax already converts almost all of it into correct
labels, so there is no headroom for a cleverer read-out to exploit. The
prior-corrected arms exceed that ceiling only because they do not read the plan
proportionally: they rescale by source class size, which is
Eq.~\ref{eq:cmm} applied after the solve rather than before.

\begin{table}[ht]
\centering
\caption{The coupling-mass ceiling, natural marginal. \emph{Coupling mass} is the
share of total transported mass landing on a below-capacity class and
\emph{expected recall} is what a perfect read-out of that plan could achieve; both
are read-out-independent properties of the coupling. The last column is what
argmax actually extracts as a fraction of that ceiling.}
\label{tab:ceiling}
\footnotesize
\begin{tabular}{lcccc}
\toprule
dataset & coupling mass & expected recall & argmax recall & extracted\\
\midrule
Jint        & $0.0041$ & $0.041$ & $0.040$ & $96\%$\\
Janesick rep1 & $0.0190$ & $0.077$ & $0.071$ & $93\%$\\
Janesick rep2 & $0.0160$ & $0.183$ & $0.211$ & $115\%$\\
\bottomrule
\end{tabular}
\end{table}

\paragraph{Which axis moves the number, and a second collapse.} Holding argmax
fixed while the marginal changes lifts below-capacity recall by $15.4\times$,
$3.7\times$ and $2.8\times$ on Jint and the two replicates, against $5.0\times$,
$2.7\times$ and $1.9\times$ for the best read-out at a fixed marginal; and the
read-out column is generous, since its best entries are the two prior-corrected
arms. Restricted to read-outs that read the plan faithfully the lift is
$1.2$ to $1.6\times$. Under the corrected marginal the read-outs also converge, from
a $5.0\times$ spread on Jint to $1.2\times$, which is the same collapse the
marginal--$\tau$ interaction shows on the $\tau$ axis: both knobs have leverage
only while the marginal is wrong. Prior correction is not free, either. On
replicate 1 it costs above-capacity recall $0.309\to0.230$ while gaining
$0.071\to0.180$ below capacity, which is why the adjudicating rule is macro-F1
with a micro floor rather than below-capacity recall alone.

\paragraph{Fitted thresholds for this battery.} On replicate 1 the median
transport-to-vote threshold ratio is $18.8\times$ for argmax and the identical soft
arms, $17.3\times$ for one-step OTP and $15.8\times$ for twenty steps, against
$0.13\times$ and $0.05\times$ for the prior-corrected arms and $1.00\times$ for the
vote by construction. Replicate 2 yields \emph{no} identified fit for any plain
read-out. These ratios rest on one to three identified fits each, with most
(slice, class) fits declined because the curve sits at chance across the entire
swept range. We therefore quote them as direction rather than magnitude; the
refusals are the more honest headline, since a read-out that never rises above
chance has no threshold to fit.

\paragraph{Marginal penalty.} Table~\ref{tab:tau} sweeps $\tau$ across five orders
of magnitude on both datasets. The ordering is monotone on each independently, the
balanced limit is the worst arm on both, and absorption tracks the starved classes
exactly. Fitting Eq.~\ref{eq:law} per arm on Janesick also shows that the threshold
is identifiable only once $\tau$ is relaxed, which is one reason
App.~\ref{app:derivation} reports no single-cell threshold ratio. Asymmetric
penalties on the two marginals were not run; we predict them at or below the balanced-marginal point, since symmetric relaxation already saturates at about half of the shape-correction effect.

\begin{table}[ht]
\centering
\caption{The $\tau$ sweep on both datasets. \emph{Identified $\rstar$} is the
number of the fourteen Janesick classes yielding a fit that is not declined; at
the default $\tau$ and tighter, every fit is declined with its ceiling pinned and
the note that the curve does not saturate within the observed window.}
\label{tab:tau}
\footnotesize
\begin{tabular}{lccccc}
\toprule
 & \multicolumn{1}{c}{Janesick} & \multicolumn{3}{c}{Jint} & \multicolumn{1}{c}{Janesick}\\
\cmidrule(r){2-2}\cmidrule(lr){3-5}\cmidrule(l){6-6}
$\tau$ & macro & macro & absorption & Endothelial / Stromal & identified $\rstar$\\
\midrule
$0.01$ (most relaxed) & $\mathbf{0.451}$ & $\mathbf{0.586}$ & $0.656$ & $0.217$ / $0.435$ & $6/14$\\
$0.1$                 & $0.399$ & $0.519$ & $0.863$ & $0.091$ / $0.222$ & $4/14$\\
$1$ (default)         & $0.306$ & $0.426$ & $0.974$ & $0.020$ / $0.058$ & $0/14$\\
$10$                  & $0.267$ & $0.407$ & $0.985$ & $0.012$ / $0.039$ & $0/14$\\
$\infty$ (balanced)   & $0.262$ & $0.404$ & $\mathbf{0.987}$ & $0.012$ / $0.036$ & $0/14$\\
\bottomrule
\end{tabular}
\end{table}

\paragraph{Marginal $\times$ $\tau$, the interaction.} Table~\ref{tab:mtau}
crosses three marginals with five $\tau$ on both domains. The design was
specified before it was run, with three pre-registered questions. The first asked
whether the $\tau$ optimum moves toward the enforced limit once the marginal is
correct. It does on the single-cell oracle row, monotonically across all five
values, and it does not on Office-Home, whose oracle row is flat; since neither
row's span is significant, we do not claim it. The leverage collapse is the sound
form of the same prediction and it holds decisively on both datasets. The second
asked whether an imperfect estimate wants some relaxation as a hedge against estimation error. It does not. The Office-Home deployable row peaks at the most relaxed $\tau$, exactly like the default row, and the single-cell deployable row peaks at $\tau=10$ by $4.5\times10^{-5}$ against a standard error of $0.072$, which is an argmax tie in a flat row rather than a sweet spot. There is no middle-ground regime on either domain. The third asked whether the combination beats the better single knob, and it does not, by $+0.0007$ and $+0.0100$ at the oracle and $+0.0270$ and $+0.0025$ at the deployable estimator.

\paragraph{Bracketing the relaxed end.} Two rows in Table~\ref{tab:mtau} peak at
$\tau=0.01$, the edge of the grid, so we extended both domains by one decade. No
$\tau{=}0.001$ minus $\tau{=}0.01$ difference is significant on either domain, the
largest reaching $z=+0.81$, and the Office-Home deployable row now has an interior
optimum at $\tau=0.01$. Rows that still peak at the new edge do so by amounts
indistinguishable from zero, so a further decade of relaxation cannot change any
conclusion. We report leverage on the pre-registered five-$\tau$ grid and treat
the extension as a bracketing check.

Absorption gives the mechanism. On single-cell data at the default marginal with
the constraint enforced, $98.0\%$ of below-capacity points are absorbed by a
covered class; the oracle marginal at the same $\tau$ cuts that to $28.4\%$.
Within the default row relaxation is what suppresses absorption, from $0.98$ to
$0.69$, whereas within the oracle row enforcement is, from $0.50$ to $0.28$. The
direction in which $\tau$ moves absorption reverses with the correctness of the
marginal, which is why a $\tau$ sweep run on a wrong marginal was always going to
look as though rigidity were the problem.

\begin{table}[ht]
\centering
\caption{Marginal $\times$ $\tau$, below-capacity macro-F1. \emph{Deployable} is
JCPOT on Office-Home and the $k$NN pre-pass on single-cell data, each the leading estimator in its own panel; \emph{oracle} uses the true composition and is an upper bound rather than a method. \emph{Span} is best minus worst within the row, with its $z$. On a wrong marginal $\tau$ is a large lever; on a corrected one no setting is distinguishable from any other. Office-Home is $4$ domain pairs $\times$ $3$ seeds and single-cell is $3$ folds $\times$ $3$ seeds, both complete. A further arm at $\tau=0.001$ leaves every conclusion unchanged (App.~\ref{app:qa}).}
\label{tab:mtau}
\footnotesize
\begin{tabular}{llccccccc}
\toprule
domain & marginal & $\tau{=}0.01$ & $0.1$ & $1$ & $10$ & $\infty$ & span & $z$\\
\midrule
\multirow{3}{*}{Office-Home}
 & default    & $\mathbf{0.450}$ & $0.359$ & $0.217$ & $0.193$ & $0.190$ & $\mathbf{0.260}$ & $11.8$\\
 & deployable & $\mathbf{0.477}$ & $0.451$ & $0.431$ & $0.426$ & $0.426$ & $0.051$ & $1.8$\\
 & oracle     & $0.528$ & $\mathbf{0.542}$ & $0.541$ & $0.538$ & $0.537$ & $0.014$ & $0.5$\\
\midrule
\multirow{3}{*}{single-cell}
 & default    & $\mathbf{0.408}$ & $0.203$ & $0.057$ & $0.044$ & $0.042$ & $\mathbf{0.366}$ & $8.7$\\
 & deployable & $0.547$ & $0.542$ & $0.552$ & $\mathbf{0.555}$ & $0.555$ & $0.013$ & $0.2$\\
 & oracle     & $0.549$ & $0.572$ & $0.634$ & $0.643$ & $\mathbf{0.644}$ & $0.096$ & $1.3$\\
\bottomrule
\end{tabular}
\end{table}

\paragraph{Entropic regularizer and detection.} Seven values of $\varepsilon$,
three folds and three seeds, on a fixed-per-class target matching the main
detector run exactly, so selectivities sit on the same scale.
Table~\ref{tab:eps} shows the Pareto block: selectivity rises by a factor of seven across the sweep while below-capacity recall collapses to zero, so the detector improves precisely as there stops being anything to detect. The confidence columns give the reason. Softening lowers confidence on correct and on incorrect points together, and expected calibration error tracks the latter exactly, so the widening gap is a rescaling rather than a separation. The sweep covers every scorable unit: one fold carries no below-capacity class and is skipped, so the denominator is smaller than the fold count elsewhere in this
paper, and the selectivity at the default $\varepsilon$ is therefore not directly comparable to the main detector suite.

\begin{table}[ht]
\centering
\caption{The $\varepsilon$ sweep, as a multiple of the calibrated default. No setting has both meaningful below-capacity recall and a detector clearing the pre-registered bar. The two smallest values produce a numerically degenerate plan. \emph{Sel.} is the selectivity of the best output-side detector; \emph{conf.} are mean confidences on correct covered points and on wrong below-capacity points. Six scorable units: one fold has no below-capacity class, so the detector suite is undefined there and it contributes no rows. The bar is cleared on at most three of six units at any $\varepsilon$, never all six.}
\label{tab:eps}
\footnotesize
\begin{tabular}{lcccccc}
\toprule
$\varepsilon$ & degenerate & below-cap.\ recall & macro & sel. & conf.\ correct & conf.\ wrong\\
\midrule
$0.01\times$ & yes & $0.000$ & $0.143$ & $1.00$ & --- & ---\\
$0.03\times$ & yes & $\mathbf{0.132}$ & $0.312$ & $1.15$ & --- & ---\\
$0.1\times$  & no  & $0.027$ & $0.441$ & $2.94$ & $0.974$ & $0.874$\\
$0.3\times$  & no  & $0.027$ & $\mathbf{0.442}$ & $2.89$ & --- & ---\\
$1\times$    & no  & $0.024$ & $0.438$ & $3.28$ & $0.927$ & $0.744$\\
$3\times$    & no  & $0.002$ & $0.408$ & $5.15$ & $0.817$ & $0.554$\\
$10\times$   & no  & $0.000$ & $0.278$ & $\mathbf{23.33}$ & $0.632$ & $0.416$\\
\bottomrule
\end{tabular}
\end{table}

\section{The decomposition under a deployable class prior}
\label{app:estpi}

Sec.~\ref{sec:decomp} defines removable and residual against the oracle marginal.
This appendix recomputes both with each deployable estimator substituted for the
oracle, on the same $780$ Office-Home class-observations and the same
rank-normalisation. No optimal transport problem was re-solved; the analysis reads
the logged per-class recalls of arms that already appear in Table~\ref{tab:fix}.
The achieved $\rho(\log r,\text{separability})=+0.108$ is a property of the design
and is therefore identical across arms.

\begin{table}[!htb]
\centering
\caption{The decomposition under seven priors, Office-Home, $n=780$. Cells are
partial Spearman $\rho$ with the other predictor controlled. Removable stays
coverage-governed and residual stays separability-governed on every arm, while the
near-zero cells at the oracle become non-zero under every estimate.}
\label{tab:estpi}
\footnotesize
\begin{tabular}{lcccc}
\toprule
prior for $\ptgt$ & removable $\sim\log r$ & removable $\sim$ sep & residual $\sim$ sep & residual $\sim\log r$\\
\midrule
oracle (published) & $\mathbf{-0.813}$ & $+0.135$ & $\mathbf{-0.338}$ & $+0.034$\\
uniform            & $-0.573$ & $+0.316$ & $-0.502$ & $-0.157$\\
$k$NN pre-pass     & $-0.517$ & $+0.263$ & $-0.374$ & $\mathbf{-0.435}$\\
JCPOT              & $-0.516$ & $+0.188$ & $-0.313$ & $-0.287$\\
BBSE               & $-0.541$ & $+0.282$ & $-0.422$ & $-0.296$\\
RLLS               & $-0.538$ & $+0.289$ & $-0.429$ & $-0.305$\\
MLLS               & $\mathbf{-0.655}$ & $+0.385$ & $-0.513$ & $\mathbf{-0.072}$\\
\bottomrule
\end{tabular}
\end{table}

\paragraph{Reading the table.} Two of the four relations are stable. Removable is
governed by coverage on every arm, at raw $\rho$ from $-0.487$ to $-0.607$ with
every arm at $p<10^{-47}$, and residual is governed by separability on every arm,
at raw $\rho$ from $-0.330$ to $-0.519$ with every arm at $p<3{\times}10^{-21}$.
The two near-zero cells at the oracle behave differently from one another. The
residual's coverage cell becomes significantly negative under every estimate,
which is the leak discussed in Sec.~\ref{sec:decomp}. The removable's separability
cell becomes significantly positive under every estimate, from $+0.108$ to
$+0.238$ raw, which is why we reserve the word dissociation for the oracle. We
also note that the oracle's own partial for this cell is $+0.135$ rather than the
raw $-0.009$, because removable is strongly driven by $\log r$ and the two
predictors retain a small positive association by construction.

\paragraph{Uniform balancing is not an estimate.} Its leak of $-0.157$ is smaller
than that of several genuine estimators, but uniform is a fixed balanced prior
rather than an estimate of $\ptgt$, and the Office-Home deep-starvation target is
close enough to balanced that the fixed prior is accidentally near-correct. Its
value would not transfer to a target with a different marginal shape. Among
genuine estimators the ordering MLLS, then JCPOT, BBSE and RLLS, then the $k$NN
pre-pass is the meaningful one.

\begin{figure}[!htb]
\centering
\includegraphics[width=0.62\textwidth]{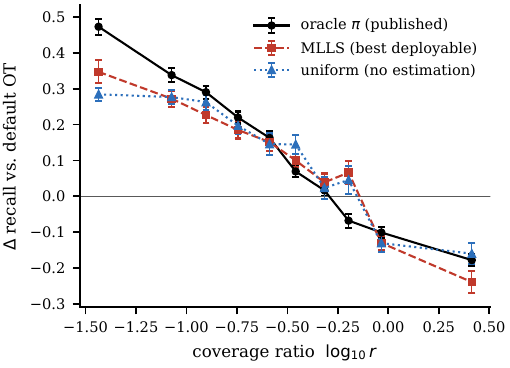}
\caption{\textbf{Dose-response under a deployable class prior.} Coverage deciles
as in Fig.~\ref{fig:dose}, with the oracle marginal, the best deployable estimator
and uniform balancing overlaid. The shape, the sign change and the cost at high
coverage survive; the amplitude at the starved end and the break-even coverage do
not.}
\label{fig:estpi}
\end{figure}

\paragraph{Dose-response, overlaid.} Fig.~\ref{fig:estpi} plots the coverage
deciles of Fig.~\ref{fig:dose} with the best deployable estimator and with uniform
balancing overlaid on the oracle. Both estimated curves keep the shape, the sign
change and the cost at high coverage. The starved end loses about a quarter of the
oracle amplitude, the break-even point moves from $r\approx0.64$ to
$r\approx0.92$, and strict monotonicity is an oracle property: the estimated curves
each contain one inversion, well inside overlapping error bars.

\paragraph{Single-cell companion.} The same analysis on the Janesick estimator
panel ($n=84$ class-observations, seven arms) is a companion rather than a
replication, since the sample is nine times smaller and the two predictors remain
correlated at $\rho(\log r,\text{separability})=-0.167$. Two of the four relations
replicate cleanly on all seven arms: removable is coverage-governed, at partial
$\rho$ from $-0.41$ to $-0.73$ with every arm at $p<10^{-4}$, and residual is
separability-governed, at partial $\rho$ from $-0.45$ to $-0.71$ with every arm at
$p<2{\times}10^{-5}$. Two things break and we report them rather than smoothing
them. First, removable depends on separability there even at the oracle
($\rho=+0.523$, $p=3{\times}10^{-7}$), so the exact orthogonality of the second leg
is specific to the constructed Office-Home design rather than a general property
of the decomposition. Second, the residual's coverage leak is not detectable on
any Janesick arm ($p$ from $0.18$ to $0.91$), which at this sample size and with
correlated predictors is an absence of evidence rather than evidence of absence.
No single-cell table logs a separability column alongside all seven arms for the
in-house cohort, so that version would require a re-run and is not reported.

\section{How close must the composition estimate be?}
\label{app:alpha}

Eq.~\ref{eq:cmm} needs the target composition, and Sec.~\ref{sec:estimation}
shows that no estimator recovers it exactly. This appendix asks a more useful
question than whether an estimate is right: how close does it have to be? We
interpolate the source marginal between the two endpoints the paper already
reports, writing
$\ptgt_\alpha=(1-\alpha)\,\ptgt^{\mathrm{s}}+\alpha\,\ptgt$ and
$\mu_j=\ptgt_\alpha[y_j]/\nsrc_{y_j}$, so that $\alpha=0$ recovers the default
marginal exactly and $\alpha=1$ recovers the oracle. Both endpoints therefore
reproduce published arms, which we use as a validation gate: the single-cell
endpoints match Table~\ref{tab:fix} to three decimals, and the Office-Home
endpoints agree to within $0.010$, inside the per-experiment composition-draw
band described in App.~\ref{app:data}. All arms use $\tau=1$, the calibrated
$\varepsilon$, and natural target composition; the relative floor of
Sec.~\ref{sec:estimation} is not applied, since both endpoints already place
positive mass on every class.

\begin{table}[!htb]
\centering
\caption{Performance along the path from the source composition to the target.
Single-cell is $3$ folds $\times$ $3$ seeds and Office-Home is $4$ domain pairs
$\times$ $3$ seeds. Rows are the eight values of $\alpha$ measured on both
domains. Absorption falls monotonically throughout, so the mechanism tracks the
marginal rather than the metric.}
\label{tab:alpha}
\footnotesize
\begin{tabular}{ccccc@{\hskip 16pt}ccc}
\toprule
& \multicolumn{3}{c}{single-cell (Jint)} & & \multicolumn{3}{c}{Office-Home}\\
\cmidrule(r){2-4}\cmidrule(l){6-8}
$\alpha$ & macro-F1 & below F1 & absorption & & macro-F1 & below F1 & absorption\\
\midrule
$0.0$ & $0.322$ & $0.062$ & $0.971$ & & $0.349$ & $0.211$ & $0.846$\\
$0.1$ & $0.383$ & $0.232$ & $0.852$ & & $0.386$ & $0.298$ & $0.754$\\
$0.2$ & $0.429$ & $0.338$ & $0.763$ & & $0.413$ & $0.368$ & $0.669$\\
$0.3$ & $0.466$ & $0.418$ & $0.685$ & & $0.434$ & $0.413$ & $0.599$\\
$0.4$ & $0.498$ & $0.476$ & $0.620$ & & $0.453$ & $0.452$ & $0.534$\\
$0.6$ & $0.548$ & $0.554$ & $0.511$ & & $0.480$ & $0.506$ & $0.417$\\
$0.8$ & $0.593$ & $0.609$ & $0.418$ & & $0.500$ & $0.532$ & $0.322$\\
$1.0$ & $\mathbf{0.623}$ & $\mathbf{0.639}$ & $0.335$ & & $\mathbf{0.510}$ & $\mathbf{0.540}$ & $0.249$\\
\bottomrule
\end{tabular}
\end{table}

\begin{figure}[!htb]
\centering
\includegraphics[width=\textwidth]{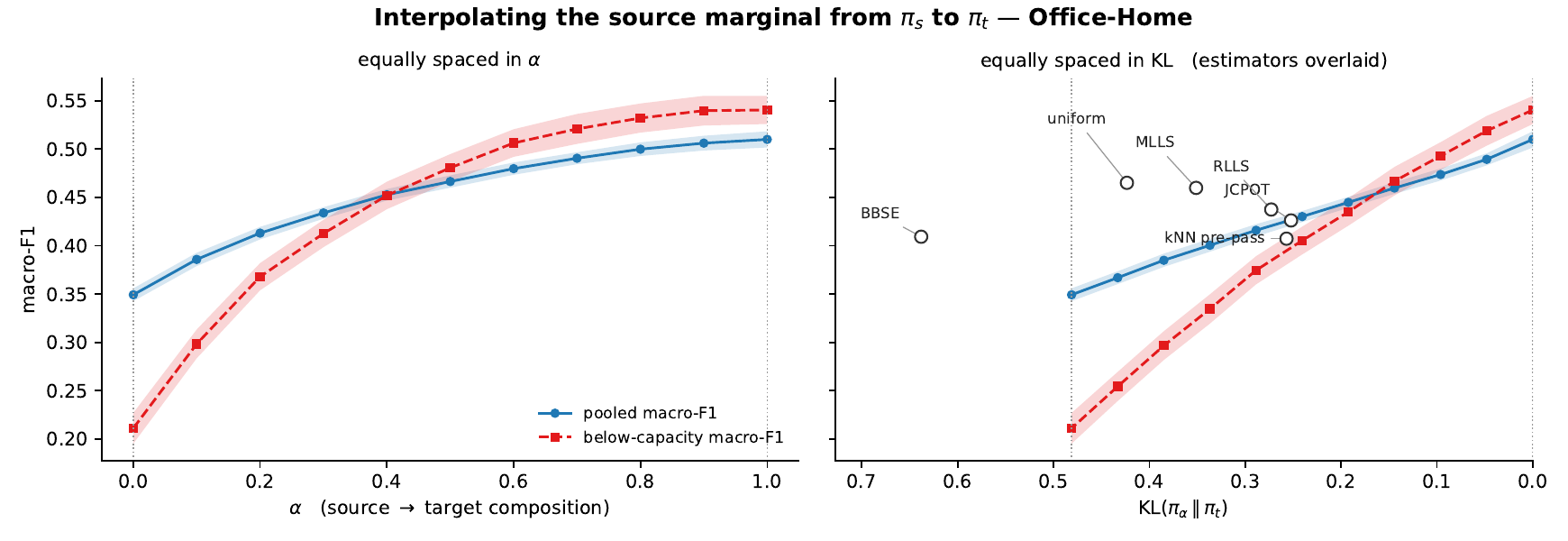}
\caption{\textbf{Office-Home.} Performance along the interpolation path, plotted
against $\alpha$ (left) and against the divergence remaining to the target
composition (right), with the divergence axis reversed so both panels read from
the default marginal to the oracle. Bands are standard errors over $12$ units.
Circles mark where each estimator of Sec.~\ref{sec:estimation} sits: its
horizontal position is the divergence of its own estimate from the target, and
its vertical position is its measured below-capacity macro-F1. The two
transport-native estimators land on the curve, so for them the divergence alone
predicts what the correction will deliver.}
\label{fig:alpha_oh}
\end{figure}

\begin{figure}[!htb]
\centering
\includegraphics[width=\textwidth]{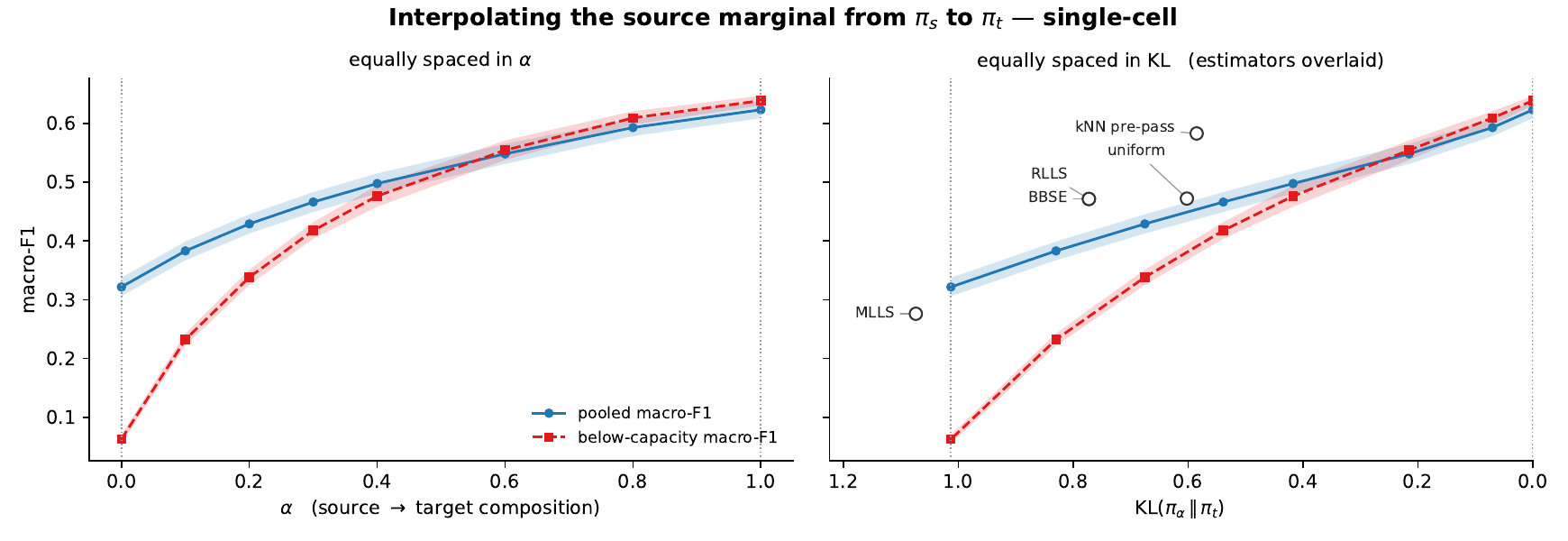}
\caption{\textbf{Single-cell.} As Fig.~\ref{fig:alpha_oh}. The horizontal axis of
the right panel spans roughly twice the range, because the single-cell
composition shift is about twice as large in divergence. The $k$NN pre-pass is
the best estimator on both axes at once: it achieves the lowest divergence to the
target composition and the highest below-capacity macro-F1 of any arm.}
\label{fig:alpha_sc}
\end{figure}

Table~\ref{tab:alpha} traces both domains along the path, and
Figs.~\ref{fig:alpha_oh} and~\ref{fig:alpha_sc} plot them.
The benefit arrives early and then flattens. Writing $\alpha_{50}$ and
$\alpha_{90}$ for the smallest $\alpha$ delivering half and nine tenths of the
total gain, both domains give $\alpha_{50}=0.30$ below capacity, and
$\alpha_{90}=0.70$ on Office-Home against $0.80$ on single-cell. An estimate only
$30\%$ of the way from the source frequencies toward the truth therefore buys half
the correction. The curve is concave on both domains, with a mean deviation above
the endpoint chord of $+0.066$ and $+0.055$: the first tenth of the path is worth
$+0.087$ and $+0.170$ below capacity, and the last tenth $0.000$ and $+0.030$.
Saturation is later on single-cell, which is what its larger composition shift
predicts, since $\mathrm{KL}(\ptgt^{\mathrm{s}}\Vert\ptgt)$ averages $1.013$ there
against $0.481$ on Office-Home. No $\alpha<1$ beats the oracle by more than one
standard error on either domain, so the path rises to its endpoint without
overshooting it.

\begin{table}[!htb]
\centering
\caption{Where each estimator of Sec.~\ref{sec:estimation} falls relative to the
path. \emph{Divergence} is the mean $\mathrm{KL}$ from the estimate to the target
composition, and \emph{path equivalent} is the $\alpha$ at which the interpolation
reaches the same divergence. Estimators whose divergence exceeds the endpoint
separation lie off the swept range and are omitted.}
\label{tab:alphaest}
\footnotesize
\begin{tabular}{llccc}
\toprule
domain & estimate of $\ptgt$ & divergence & path equivalent & below-capacity F1\\
\midrule
\multirow{5}{*}{Office-Home}
 & JCPOT          & $0.252$ & $0.264$ & $0.426$\\
 & $k$NN pre-pass & $0.257$ & $0.258$ & $0.407$\\
 & RLLS           & $0.273$ & $0.234$ & $0.437$\\
 & MLLS           & $0.351$ & $0.236$ & $0.460$\\
 & uniform        & $0.423$ & $0.071$ & $0.465$\\
\midrule
\multirow{4}{*}{single-cell}
 & $k$NN pre-pass & $\mathbf{0.585}$ & $0.408$ & $\mathbf{0.583}$\\
 & uniform        & $0.602$ & $0.317$ & $0.472$\\
 & BBSE           & $0.773$ & $0.140$ & $0.471$\\
 & RLLS           & $0.773$ & $0.140$ & $0.471$\\
\bottomrule
\end{tabular}
\end{table}

Placing the estimators on the same axes turns the path into a diagnostic
(Table~\ref{tab:alphaest}). Every estimator lands on or above the curve, never
below it, so none performs worse than its divergence would predict. On
Office-Home the two transport-native estimators land on the curve within a
standard error, which means the divergence to the target composition is by itself
enough to say what the correction will deliver. Elsewhere estimators sit above
it, and they do so because divergence is not directional: mass placed on rare
classes is nearly free below capacity, so an estimate that errs toward the rare
classes costs far less than its divergence suggests. Uniform balancing is the
clean illustration on Office-Home, reaching a below-capacity macro-F1 the path
does not attain until $\alpha\approx0.45$ from a divergence equivalent to
$\alpha=0.071$. The single-cell $k$NN pre-pass is the strongest case of all: it
is simultaneously the closest estimate to the target composition and the best
performer of any arm, which is why Sec.~\ref{sec:estimation} finds it the leading
deployable estimator there.

\section{Additional results}
\label{app:additional}

\begin{table}[ht]
\centering
\caption{Solver invariance and output-side detectors, single-cell (Jint).
\emph{Left:} four independent entropic-OT implementations on identical
embeddings, points, labels, folds and seeds, with a $k$NN vote on the same
embedding for reference. \emph{Absorption} is the fraction of a starved class's target mass received by a single other class. Fixed-per-class target composition, so this is a
within-run contrast across solvers rather than a cross-design comparison; the
corrected natural-composition comparison is Table~\ref{tab:fair}. \emph{Right:}
detectors for below-capacity points, $9$ folds $\times$ $3$ seeds, against a bar
fixed in advance of selectivity at least $3\times$ at coverage at least $90\%$;
suite verdict $0/5$, with $10/135$ unit evaluations passing.
$^{\ast}$over the units where it fires.}
\label{tab:appendix}
\footnotesize
\begin{tabular}{lccc@{\hskip 16pt}lc}
\toprule
solver & macro & Endothelial & absorption & detector (side) & median sel.\ / pass\\
\midrule
$k$NN vote (reference) & $0.638$ & $0.295$ & $0.541$ & plan confidence (out) & $2.43$ / $6/27$\\
UOT $+$ centering      & $0.426$ & $0.020$ & $0.974$ & split conformal (out) & $2.38$ / $1/27$\\
moscot                 & $0.424$ & $0.014$ & $0.981$ & reject-sink OT (out)  & $1.51$ / $0/27$\\
UOT, no centering      & $0.320$ & $0.009$ & $0.994$ & transport entropy (out) & $0.42$ / $0/27$\\
                       &         &         &         & coverage rule (source)& $24.50^{\ast}$ / $3/27$\\
\bottomrule
\end{tabular}
\end{table}

\begin{figure}[ht]
\centering
\includegraphics[width=0.80\textwidth]{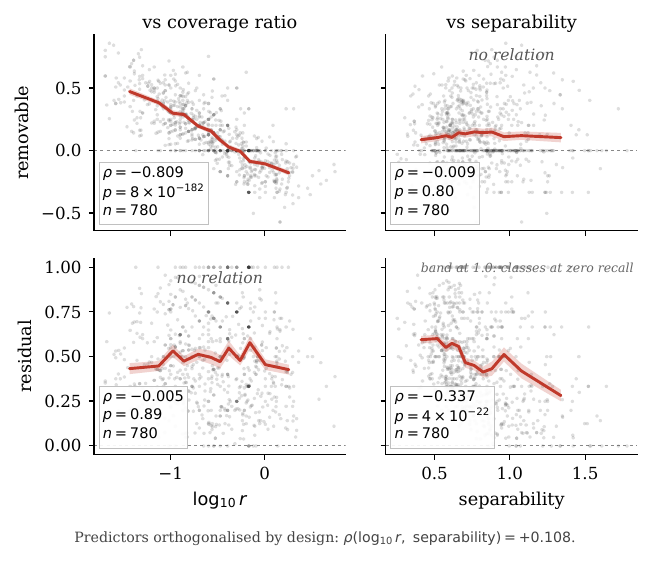}
\caption{\textbf{The double dissociation.} What the composition-matched marginal
removes (top) and what survives it (bottom), against coverage (left) and class
separability (right), on Office-Home. The two null panels are shown deliberately:
what is removed is explained by coverage and \emph{not} by separability, and what
survives is explained by separability and \emph{not} by coverage. Points are
class-observations ($4$ domain pairs $\times$ $3$ seeds), the line is a binned
mean with standard error, and the band at residual $=1.0$ is classes the
correction leaves at zero recall. The panels are interpretable only because the
predictors were orthogonalized by construction; in the single-cell data they are
confounded at $\rho=-0.618$.}
\label{fig:diss}
\end{figure}

\begin{figure}[ht]
\centering
\includegraphics[width=0.98\textwidth]{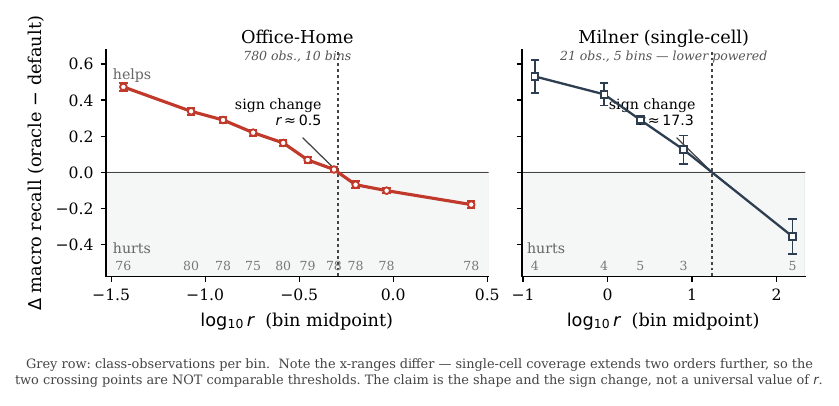}
\caption{\textbf{Dose-response of the correction.} Change in macro recall from
the composition-matched marginal against coverage, binned. Monotone on both
domains, crossing zero and turning negative on the best-covered classes. The
single-cell panel has $21$ class-observations in $5$ bins against Office-Home's
$780$ in $10$ and is a companion rather than an equally powered replication, and
the $x$-ranges differ, so the two crossing points are not comparable thresholds.}
\label{fig:dose}
\end{figure}

\paragraph{Specificity on data with no composition shift.} SpatialCorpus splits one file into source and target, so the two compositions agree by construction and a capacity correction should do nothing. It does nothing: uniform balancing leaves macro recall flat on all five tissues and lowers micro recall on every one. That is the coarse-grained counterpart of the sign change in Fig.~\ref{fig:dose}, which shows the same specificity at class resolution.

\paragraph{Backbone invariance and cut-point sensitivity.} Table~\ref{tab:robust}
collects two robustness checks. On the left, repeating the mechanism analysis on a
PCA basis reproduces every count, and PCA is the better read-out of the two, which
forecloses the reading that PCA is simply worse; on PCA the source-side coverage
rule fires in $24$ of $27$ units and passes $23$, while every output-side statistic
still fails. The same holds on Office-Home. Repeating the mechanism analysis on
ResNet-50 followed by PCA gives a capacity partial of $+0.347$ against CLIP's
$+0.329$, a distance partial of $-0.191$ against $-0.206$, and argmax counts of
$20.2\%$ nearest and $19.2\%$ largest against $18.6\%$ and $20.5\%$
($13{,}312$ pairs over $12$ units). Both counts sit near the $1$ in $7$ a chance
absorber would give, on both backbones, which is the same degeneracy the
single-cell analysis shows and a second demonstration that the partial is the
statistic that transfers. On the right, the below-capacity knob ordering of
Table~\ref{tab:knobs} holds at every cut point we tested, so the $r<0.15$ boundary
is a reporting convenience rather than a fitted one.

\begin{table}[ht]
\centering
\caption{\emph{Left:} backbone invariance, $9$ folds $\times$ $3$ seeds each on a
$40{,}000$-point source, fixed-per-class target, so a within-run contrast. Counts differ slightly from Sec.~\ref{sec:mechanism} because that run uses natural composition and this one does not.
\emph{Right:} the below-capacity knob ordering of Table~\ref{tab:knobs} at four
cut points, as $\Delta$ macro recall against the default marginal.}
\label{tab:robust}
\footnotesize
\begin{tabular}{lcc@{\hskip 20pt}lcccc}
\toprule
& foundation & PCA & cut & $n$ & shape & $\tau$ & $\varepsilon$\\
\midrule
confident errors    & $33/33$ & $33/33$ & $r<0.05$ & $56$  & $0.472$ & $0.243$ & $0.006$\\
absorber = nearest  & $0/33$  & $0/33$  & $r<0.10$ & $136$ & $0.419$ & $0.234$ & $0.014$\\
absorber = largest  & $30/33$ & $33/33$ & $r<0.15$ & $235$ & $0.366$ & $0.208$ & $0.013$\\
confidence ratio    & $0.874$ & $0.906$ & $r<0.30$ & $389$ & $0.296$ & $0.176$ & $0.015$\\
macro recall        & $0.445$ & $0.495$ &          &       &         &         & \\
detector suite      & $0/5$   & $0/5$   &          &       &         &         & \\
\bottomrule
\end{tabular}
\end{table}

\paragraph{Ratio, not sample size.} A quantity stated as a ratio invites the
objection that it is small-sample statistics in disguise. A trained classifier
separates the two. In a fixed-per-class run, the most ratio-starved class
($r=0.043$) has $117$ effective training points and a trained latent-variable
classifier \citep{xu2021scanvi} recovers it at $0.960$ while transport loses it
entirely, whereas a class four times better covered by ratio ($r=0.439$) has only
$54$ training points and that classifier loses it, at $0.122$. Transport follows
the ratio and a trained classifier follows the count.

\paragraph{Pooled and stratified readings disagree.} MLLS closes only $4.7\%$ of
the single-cell micro gap, the worst of the panel, while scoring $0.672$ below
capacity, above even the oracle's $0.546$. That is possible because the oracle
matches the \emph{whole} composition rather than optimising the starved classes,
so an estimator that over-allocates mass to rare classes can exceed it there and
pay for it pooled, which is what MLLS does. Either number alone misleads, which is
why every table in this paper carries both.

\paragraph{Estimator error is amplified, not merely passed through.} A
construction makes the failure mode concrete. Under a small \emph{absolute} floor
on estimated shares, a class the pre-pass never predicts holds share $0.0036$
against $0.0766$ for the rest, and its recall then stays exactly $0.000$ even when
its source points are a $20\times$ better match to the target points in question,
because the marginal starves the class, so it is never predicted, so the estimate
is never contradicted. A \emph{relative} floor of $0.5/K$ breaks the loop and is
what every estimator arm reported here uses. The construction is illustrative
rather than a measured arm, but the failure it describes is what we observed
before the floor was changed, and the change is a one-line design rule for
deploying Eq.~\ref{eq:cmm}: floor the estimate relative to uniform, not absolutely.

\begin{table}[ht]
\centering
\caption{Matched corrections, single-cell (Jint), $3$ folds $\times$ $3$ seeds,
natural composition. \emph{Composition-matched} arms, transport and vote alike,
use the \emph{true} target composition and are therefore matched oracles;
\emph{prior-corrected} arms divide by the source class prior and use no target
information, so they are matched to each other at a lower information level. The
oracle-to-oracle comparison is transport $0.546$ against vote $0.414$ below
capacity, where \emph{below} is the below-capacity stratum, three (fold, class)
cells averaged over seeds.}
\label{tab:fair}
\footnotesize
\begin{tabular}{lcccc}
\toprule
read-out & macro & macro-F1 & micro & below capacity\\
\midrule
transport, default        & $0.433$ & $0.322$ & $0.393$ & $0.030$\\
$k$NN vote, uncorrected   & $0.627$ & $0.553$ & $0.625$ & $0.351$\\
linear probe, uncorrected & $0.268$ & $0.212$ & $0.215$ & $0.014$\\
\midrule
$k$NN, prior-corrected    & $\mathbf{0.654}$ & $0.571$ & $0.640$ & $0.430$\\
$k$NN, composition-matched& $0.635$ & $0.577$ & $0.652$ & $0.414$\\
linear probe, corrected   & $0.452$ & $0.391$ & $0.377$ & $0.278$\\
\textbf{transport, composition-matched} & $0.619$ & $\mathbf{0.623}$ & $\mathbf{0.718}$ & $\mathbf{0.546}$\\
\bottomrule
\end{tabular}
\end{table}

\paragraph{Office-Home matched corrections, full.} Table~\ref{tab:ohfull} is the
Office-Home counterpart of Table~\ref{tab:fair}. Two caveats travel with it. The
linear-probe arms come from a separate Office-Home experiment with an independent
composition draw, so differences of this size sit near what the comparison
resolves; and the pooled metrics are dominated by well-covered classes, where
Sec.~\ref{sec:fix} shows the marginal correction helps least and can hurt.

\begin{table}[ht]
\centering
\caption{Matched corrections on Office-Home, $4$ domain pairs $\times$ $3$ seeds,
deep-starvation configuration. \emph{Below} is the below-capacity stratum,
$n=235$ class-observations.}
\label{tab:ohfull}
\footnotesize
\begin{tabular}{lcccc}
\toprule
read-out & macro & macro-F1 & micro & below capacity\\
\midrule
transport, default           & $0.388$ & $0.356$ & $0.440$ & $0.136$\\
$k$NN vote, uncorrected      & $0.367$ & $0.327$ & $0.506$ & $0.154$\\
linear probe, uncorrected    & $0.199$ & $0.136$ & $0.313$ & $0.004$\\
\midrule
$k$NN, prior-corrected       & $0.553$ & $0.483$ & $0.580$ & $0.526$\\
$k$NN, composition-matched   & $0.460$ & $0.461$ & $\mathbf{0.644}$ & $\mathbf{0.566}$\\
linear probe, class-balanced & $\mathbf{0.590}$ & $\mathbf{0.533}$ & $0.641$ & $0.542$\\
linear probe, prior-corrected& $0.389$ & $0.329$ & $0.517$ & $0.182$\\
transport, composition-matched & $0.521$ & $0.523$ & $0.623$ & $0.547$\\
\bottomrule
\end{tabular}
\end{table}

\paragraph{Which correction a simple read-out needs is domain-dependent.} An
uncorrected linear probe is catastrophic below capacity on both domains, at
$0.014$ and $0.004$, and both corrections we tested lift it, but they rank
oppositely. Dividing the posterior by the source prior reaches $0.278$ on
single-cell data where class-balanced loss re-weighting reaches $0.033$, and on
Office-Home the ordering reverses, at $0.542$ against $0.182$. We report both per
domain and make no general recommendation. An uncorrected baseline understates a
linear probe severely, which is a caution for any comparison of this kind,
including those drawn here.

\paragraph{Janesick, natural composition.} Table~\ref{tab:jan} gives macro and
micro recall by marginal on both replicates. The pattern of Table~\ref{tab:fix}
holds: uniform balancing takes the highest macro on rep2 while micro collapses,
and only the composition-matched marginal raises macro without reducing micro on
either replicate.

\begin{table}[ht]
\centering
\caption{Janesick under natural target composition, macro\,/\,micro recall by
marginal, two replicates $\times$ $3$ seeds.}
\label{tab:jan}
\footnotesize
\begin{tabular}{lcccc}
\toprule
& \multicolumn{2}{c}{replicate 1} & \multicolumn{2}{c}{replicate 2}\\
\cmidrule(r){2-3}\cmidrule(l){4-5}
estimate of $\ptgt$ & macro & micro & macro & micro\\
\midrule
source frequencies & $0.141$ & $0.224$ & $0.334$ & $0.477$\\
uniform            & $\mathbf{0.206}$ & $0.149$ & $\mathbf{0.591}$ & $0.346$\\
$k$NN pre-pass     & $0.180$ & $0.167$ & $0.479$ & $0.437$\\
JCPOT              & $0.179$ & $0.157$ & $0.461$ & $0.389$\\
BBSE               & $0.161$ & $\mathbf{0.316}$ & $0.346$ & $0.436$\\
RLLS               & $0.160$ & $0.317$ & $0.347$ & $0.436$\\
MLLS               & $0.160$ & $0.317$ & $0.345$ & $0.436$\\
\midrule
oracle $\ptgt$     & $0.204$ & $0.314$ & $0.449$ & $\mathbf{0.554}$\\
\bottomrule
\end{tabular}
\end{table}

\paragraph{Per-pair replication, Office-Home.} Oracle micro gap by domain pair:
$+0.127$, $+0.174$, $+0.198$ and $+0.185$. It is present on all four, with the
best estimator closing $53.7$ to $70.0\%$ on every one, so the estimator contrast
is not driven by a single pair. Which estimator is best is unstable across pairs,
with MLLS at $70.0\%$ on one and $15.0\%$ on another, which is why we quote the
range.

\paragraph{Minority synthesis does not add to the correction.} Within-run
contrasts on a fixed-per-class target. Synthesizing minority source points by
latent-space interpolation \citep{chawla2002smote,dablain2022deepsmote}, dosed to
the fitted threshold, adds $+0.006$ macro recall on Jint and $+0.036$ on
Janesick on its own, and $+0.000$ and $+0.002$ respectively on top of a corrected
marginal, with absorption unchanged to three decimals. We also tested the one
regime where it might have a niche, namely classes too small to reweight, with
$17$ and $24$ target points, where it does not appear.

\paragraph{Impossibility cases.} One target slide contains two classes with
\emph{zero} source points. These are absent rather than under-covered, since no
marginal can place mass that does not exist, and they are excluded from all
below-capacity means.

\newpage
\section*{NeurIPS Paper Checklist}

\begin{enumerate}

\item {\bf Claims}
    \item[] Question: Do the main claims made in the abstract and introduction accurately reflect the paper's contributions and scope?
    \item[] Answer: \answerYes{}
    \item[] Justification: The abstract and Sec.~\ref{sec:intro} state the capacity mechanism, the limits of downstream repair and the restricted operating range of the correction, each supported by Secs.~\ref{sec:capacity} and~\ref{sec:correction}.

\item {\bf Limitations}
    \item[] Question: Does the paper discuss the limitations of the work performed by the authors?
    \item[] Answer: \answerYes{}
    \item[] Justification: Sec.~\ref{sec:discussion} contains a dedicated Limitations paragraph covering the oracle marginal, the coupling of the two components under estimated compositions, the small below-capacity stratum, the scope of the parametric law and the protocol exceptions.

\item {\bf Theory assumptions and proofs}
    \item[] Question: For each theoretical result, does the paper provide the full set of assumptions and a complete (and correct) proof?
    \item[] Answer: \answerYes{}
    \item[] Justification: App.~\ref{app:derivation} derives Eq.~\ref{eq:law} with assumptions and scope.

    \item {\bf Experimental result reproducibility}
    \item[] Question: Does the paper fully disclose all the information needed to reproduce the main experimental results of the paper to the extent that it affects the main claims and/or conclusions of the paper (regardless of whether the code and data are provided or not)?
    \item[] Answer: \answerYes{}
    \item[] Justification: App.~\ref{app:data} gives datasets, folds, source budgets, solver settings and metric definitions, and Sec.~\ref{sec:capacity} states the evaluation protocol and the pre-registered adjudication rule.

\item {\bf Open access to data and code}
    \item[] Question: Does the paper provide open access to the data and code, with sufficient instructions to faithfully reproduce the main experimental results, as described in supplemental material?
    \item[] Answer: \answerYes{}
    \item[] Justification: Janesick (GSE243275), SpatialCorpus and Office-Home are public, and code and configurations for every arm will be released after the main conference acceptance. The main results can be reproduced with the technical details in the paper. The in-house cohort cannot be released yet.

\item {\bf Experimental setting/details}
    \item[] Question: Does the paper specify all the training and test details (e.g., data splits, hyperparameters, how they were chosen, type of optimizer) necessary to understand the results?
    \item[] Answer: \answerYes{}
    \item[] Justification: Secs.~\ref{sec:capacity} and App.~\ref{app:data} specify the encoder, cost, marginals, solver, folds, seeds, source budgets and target-composition design for every arm.

\item {\bf Experiment statistical significance}
    \item[] Question: Does the paper report error bars suitably and correctly defined or other appropriate information about the statistical significance of the experiments?
    \item[] Answer: \answerYes{}
    \item[] Justification: Results are reported over multiple folds, replicates and seeds; correlations carry $p$-values (Secs.~\ref{sec:mechanism} and~\ref{sec:decomp}), knob comparisons carry $z$-statistics (Sec.~\ref{sec:knobs}, App.~\ref{app:qa}) and binned curves carry standard errors (Apps.~\ref{app:alpha} and~\ref{app:additional}).

\item {\bf Experiments compute resources}
    \item[] Question: For each experiment, does the paper provide sufficient information on the computer resources (type of compute workers, memory, time of execution) needed to reproduce the experiments?
    \item[] Answer: \answerNo{}
    \item[] Justification: Wall-clock and hardware budgets are not reported. The experiments are transport solves and re-analyses of logged per-class recalls rather than model training, and App.~\ref{app:data} gives the solver settings and convergence criteria.
    
\item {\bf Code of ethics}
    \item[] Question: Does the research conducted in the paper conform, in every respect, with the NeurIPS Code of Ethics \url{https://neurips.cc/public/EthicsGuidelines}?
    \item[] Answer: \answerYes{}
    \item[] Justification: The research conforms with the NeurIPS Code of Ethics.

\item {\bf Broader impacts}
    \item[] Question: Does the paper discuss both potential positive societal impacts and negative societal impacts of the work performed?
    \item[] Answer: \answerYes{}
    \item[] Justification: The work clarifies when an optimal-transport read-out can be trusted and how to correct it, and we do not anticipate direct negative societal impacts from a methodological correction of this kind.
    
\item {\bf Safeguards}
    \item[] Question: Does the paper describe safeguards that have been put in place for responsible release of data or models that have a high risk for misuse (e.g., pre-trained language models, image generators, or scraped datasets)?
    \item[] Answer: \answerNA{}
    \item[] Justification: The paper releases no pretrained models, generative systems or scraped data, so it poses no such risk of misuse.

\item {\bf Licenses for existing assets}
    \item[] Question: Are the creators or original owners of assets (e.g., code, data, models), used in the paper, properly credited and are the license and terms of use explicitly mentioned and properly respected?
    \item[] Answer: \answerYes{}
    \item[] Justification: Public datasets and software used are cited in Sec.~\ref{sec:capacity} and App.~\ref{app:data}, with the accession number given for the public spatial cohort.

\item {\bf New assets}
    \item[] Question: Are new assets introduced in the paper well documented and is the documentation provided alongside the assets?
    \item[] Answer: \answerNA{}
    \item[] Justification: No new datasets or models are introduced.

\item {\bf Crowdsourcing and research with human subjects}
    \item[] Question: For crowdsourcing experiments and research with human subjects, does the paper include the full text of instructions given to participants and screenshots, if applicable, as well as details about compensation (if any)? 
    \item[] Answer: \answerNA{}
    \item[] Justification: The paper involves no crowdsourcing and no experiments with human subjects.

\item {\bf Institutional review board (IRB) approvals or equivalent for research with human subjects}
    \item[] Question: Does the paper describe potential risks incurred by study participants, whether such risks were disclosed to the subjects, and whether Institutional Review Board (IRB) approvals (or an equivalent approval/review based on the requirements of your country or institution) were obtained?
    \item[] Answer: \answerNA{}
    \item[] Justification: The study is a secondary analysis of existing single-cell, spatial transcriptomics and image benchmarks, and involves no human-subjects research conducted by the authors.

\item {\bf Declaration of LLM usage}
    \item[] Question: Does the paper describe the usage of LLMs if it is an important, original, or non-standard component of the core methods in this research? Note that if the LLM is used only for writing, editing, or formatting purposes and does \emph{not} impact the core methodology, scientific rigor, or originality of the research, declaration is not required.
    \item[] Answer: \answerNA{}
    \item[] Justification: LLMs were used only for writing and editing the paper and are not part of the core methodology.

\end{enumerate}

\end{document}